\def\year{2016}\relax
\documentclass[letterpaper]{article}
\pdfoutput=1
\usepackage{aaai16}
\usepackage{times}
\usepackage{helvet}
\usepackage{courier}

\usepackage{amsmath,amssymb}
\usepackage{soul,color}
\usepackage{enumerate}
\usepackage{algorithm}
\usepackage{algorithmic}
\usepackage{graphicx}
\usepackage{subcaption}
\usepackage{xcolor,colortbl}
\usepackage{xspace}
\usepackage{tabularx}
\usepackage{multirow}
\usepackage{mathtools}
\usepackage{amsthm}
\usepackage{booktabs}

\usepackage{etoolbox}
\newtoggle{fullpaper}
\toggletrue{fullpaper} % uncomment to get arXiv version
\usepackage{xr}
\renewcommand{\algorithmicrequire}{\textbf{Input:}}
\renewcommand{\algorithmicensure}{\textbf{Output:}}

\theoremstyle{plain}
\newtheorem{thm}{Theorem} % reset theorem numbering for each chapter
\theoremstyle{definition}
\newtheorem{defn}[thm]{Definition} % definition numbers are dependent on theorem numbers
\newtheorem{lemma}{Lemma}

\DeclareMathOperator*{\argmin}{argmin}
\DeclarePairedDelimiter{\floor}{\lfloor}{\rfloor}

\newcommand{\vbf}[1]{\ensuremath{\boldsymbol{\mathrm{#1}}}}
\newcommand{\norm}[1]{\ensuremath{\lVert{#1}\rVert}_{F}^{2}}

\newcommand{\eg}[0]{\emph{e.g.},\xspace}
\newcommand{\ie}[0]{\emph{i.e.},\xspace}
\renewcommand{\st}[0]{\ensuremath{\mbox{s.t.\xspace}}}

\newcommand{\instA}[0]{\ensuremath{^{\dagger}}}
\newcommand{\instB}[0]{\ensuremath{^{\star}}}
\newcommand{\instC}[0]{\ensuremath{^{\ddagger}}}
\newcommand{\instD}[0]{\ensuremath{^{\S}}}
\newcommand{\citet}[1]{\citeauthor{#1}~(\citeyear{#1})\xspace}

\newcolumntype{a}{>{\columncolor{green!25}}c}

\begin{document}
% The file aaai.sty is the style file for AAAI Press 
% proceedings, working notes, and technical reports.
%
\title{MOOCs Meet Measurement Theory: A Topic-Modelling Approach}
%\author{Jiazhen He\instA\instB \hspace{1em} Benjamin I. P. Rubinstein\instA \hspace{1em}
%James Bailey\instA\instB \hspace{1em} Rui Zhang\instA\\
%\instA Dept. Computing and Information Systems, The University of Melbourne, Australia
%\instB National ICT Australia\\
%jiazhenh@student.unimelb.edu.au,
%\{baileyj, brubinstein, rui.zhang\}@unimelb.edu.au,\\
%\vspace{-1em}
%\AND
%Sandra Milligan\instC \hspace{1em}
%Jeffrey Chan\instD
%\\
%\instC Melbourne Graduate School of Education, The University of Melbourne, Australia \\
%\instD School of Computer Science and Information Technology, 
%RMIT University \\
%s.milligan@unimelb.edu.au, jeffrey.chan@rmit.edu.au
%}

\author{Jiazhen He\instA\instB \hspace{1em} Benjamin I. P. Rubinstein\instA \hspace{1em}
	James Bailey\instA\instB \hspace{1em} Rui Zhang\instA \\
	\textbf {\Large Sandra Milligan\instC \hspace{1em}
	Jeffrey Chan\instD}
    \\
    \instA Dept. Computing and Information Systems, The University of Melbourne, Australia 
    \ \ \ \instB National ICT Australia\\
    \instC Melbourne Graduate School of Education, The University of Melbourne, Australia \\
    \instD School of Computer Science and Information Technology, 
    RMIT University, Australia \\
    \{jiazhenh@student., baileyj, brubinstein, rui.zhang, s.milligan\}@unimelb.edu.au,
    jeffrey.chan@rmit.edu.au
}

\maketitle
\begin{abstract}
\begin{quote}
This paper adapts topic models to the psychometric testing of MOOC students based on their online forum postings. Measurement theory from education and psychology provides statistical models for quantifying a person's attainment of intangible attributes such as attitudes, abilities or intelligence. Such models infer latent skill levels by relating them to individuals' observed responses on a series of items such as quiz questions. 
The set of items can be used to measure a latent skill if individuals' responses on them conform to a Guttman scale. Such well-scaled items differentiate between individuals and inferred levels span the entire range from most basic to the advanced. 
In practice, education researchers manually devise items (quiz questions) while optimising well-scaled conformance. Due to the costly nature and expert requirements of this process, psychometric testing has found limited use in everyday teaching. 
We aim to develop usable measurement models for highly-instrumented MOOC delivery platforms, by using participation in automatically-extracted online forum topics as items.
The challenge is to formalise the Guttman scale educational constraint and incorporate it into topic models.
To favour topics that automatically conform to a Guttman scale, we introduce a novel regularisation into non-negative matrix factorisation-based topic modelling. We demonstrate the suitability of our approach with both quantitative experiments on three Coursera MOOCs, and with a qualitative survey of topic interpretability on two MOOCs by domain expert interviews.
\end{quote}
\end{abstract}

\section{Introduction}
Massive Open Online Courses (MOOCs) have recently been the subject of a number of studies within disciplines as varied as education, psychology and computer science~\cite{ramesh2014uncovering,anderson2014engaging,kizilcec2013deconstructing,diez2013peer,milligan2015crowd}. With few studies taking a truly cross-disciplinary approach, this paper is the first to marry topic modelling with measurement theory from education and psychology.

\textit{Measurement} in education and psychology is the process of assigning a number to an attribute of an individual in such a way that individuals can be compared to one another~\cite{pedhazur1991measurement}. These attributes are often intangible 
such as attitudes, abilities or intelligence.
Since the attribute to be measured is not directly observable, a set of items is often devised manually and individuals' responses on the items are collected. 
Based on a modelled correspondence with observed item responses, latent attribute levels of a cohort can be inferred. This process is called \emph{scaling}~\cite{de2013theory}. 

A \textit{Guttman scale}~\cite{guttman1950basis} is one which
induces a total ordering 
on items---an individual who successfully answers/agrees with a particular item also answers/agrees with items of lower rank-order.
Table~\ref{tab:guttman} depicts an example Guttman scale measuring mathematical ability~\cite{abdi2010guttman}, where the items are ordered in increasing latent difficulty, from \emph{Counting} to \emph{Division}. Here the total score corresponds to the persons' latent ability: the greater the higher.

\begin{table}[htbp]
    \centering
    \scriptsize
    \caption{An example of a perfect Guttman scale measuring mathematical ability\cite{abdi2010guttman} , where 1 means the person has mastered the item and 0 for not. Person 5 who has mastered the most difficult item \emph{Division}, is expected to have mastered all easier items as well.}
    \begin{tabular}{|a|ccccc|c|}
        \hline
        \rowcolor{green!25}
        & \textbf{Item 1}& \textbf{Item 2} & \textbf{Item 3} & \textbf{Item 4} & \textbf{Item 5} & \textbf{Total} \\
        \rowcolor{green!25}
        & \textbf{(Counting)}  & \textbf{($+$)} & \textbf{($-$)} & \textbf{($\times$)} & \textbf{($\div$)} & \textbf{Score} \\
        \hline
        
        \textbf{Person 1} & 1 \cellcolor{red!25}     & 0 \cellcolor{blue!35}     & 0 \cellcolor{blue!35}    & 0 \cellcolor{blue!35}     & 0 \cellcolor{blue!35}    & 1 \cellcolor{green!25} \\
        \textbf{Person 2} & 1 \cellcolor{red!25}     & 1  \cellcolor{red!25}    & 0 \cellcolor{blue!35}     & 0  \cellcolor{blue!35}    & 0 \cellcolor{blue!35}    & 2 \cellcolor{green!25} \\
        \textbf{Person 3} & 1  \cellcolor{red!25}   & 1 \cellcolor{red!25}    & 1 \cellcolor{red!25}    & 0   \cellcolor{blue!35}  & 0  \cellcolor{blue!35}   & 3 \cellcolor{green!25} \\
        \textbf{Person 4} & 1 \cellcolor{red!25}   & 1  \cellcolor{red!25}   & 1   \cellcolor{red!25}  & 1  \cellcolor{red!25}   & 0 \cellcolor{blue!35}    & 4 \cellcolor{green!25} \\
        \textbf{Person 5} & 1 \cellcolor{red!25}   & 1 \cellcolor{red!25}    & 1  \cellcolor{red!25}   & 1  \cellcolor{red!25}   & 1  \cellcolor{red!25}    & 5 \cellcolor{green!25} \\
        \hline
    \end{tabular}%	
    \label{tab:guttman}%
\end{table}%

In MOOCs, as in the traditional classroom, we may hypothesise that students possess a latent ability in the subject at hand. For example, in a MOOC on macroeconomics, students are expected to develop knowledge in introductory macroeconomics via videos, quizzes and forums. Students' latent abilities can be defined, validated and measured using indicators drawn from student responses to activities like interaction with videos, quiz results and forum participation. Unlike the traditional classroom, MOOCs create new challenges and opportunities for measurement through the multiple modes of student interaction online---all monitored at large scale. The education research community is broadly interested in whether and how latent complex patterns of engagement might evidence the possession of a latent skill, and not just explanatory variables (\eg visible quizzes and assignments) by themselves~\cite{milligan2015crowd}.

This paper focuses on using the content of forum discussion in MOOCs for measurement, which is too time-consuming to analyse manually but that can provide a predictive indicator of achievement~\cite{Beaudoin2002147}.
We automatically generate items (topics) from unstructured forum data using topic modelling. 
Our goal is to discover items on which dichotomous (posting on a topic or not) student responses conform to a Guttman scale; where items are interpretable to subject-matter experts who could be teaching such MOOCs.
For example, for a MOOC on discrete optimisation, our goal is to automatically discover topics such as \emph{How to use platform/python}---the easiest which most students contribute to---and \emph{How to design and tune simulated annealing and local search}---a more difficult topic which only a few students might post on.
Such well-scaled items can be used for curriculum design and student assessment.

The challenge is to formalise the Guttman scale educational constraint and incorporate it into topic models.
We opt to focus on non-negative matrix factorisation (NMF) approaches to topic modelling, as these admit natural integration of the Guttman scale educational constraint. 
\paragraph{Contributions.}
The main contributions of this paper are:
\begin{itemize}
    \item A first study of how a machine learning technique, NMF-based topic modelling, can be used for the education research topic of psychometric testing;
    \item A novel regularisation of NMF that incorporates the educational constraint that inferred topics form a Guttman scale; and accompanying training algorithm;
    \item Quantitative experiments on three Coursera MOOCs covering a broad swath of disciplines, establishing statistical effectiveness of our algorithm; and
    \item A carefully designed qualitative survey of experts in two MOOC subjects, which supports the interpretability of our results and suggests their applicability in education.
\end{itemize}

\section{Related Work}
Various studies have been conducted 
into MOOCs for tasks such as dropout prediction~\cite{halawa2014dropout,yang2013turn,ramesh2014learning,kloft2014predicting,he2015identifying}, characterising student engagement~\cite{anderson2014engaging,kizilcec2013deconstructing,ramesh2014learning} and peer assessment~\cite{diez2013peer,piech2013tuned,mi2015probabilistic}.

Forum discussions in MOOCs have been of interest recently, due to the availability of rich textual data and social behaviour.
For example, \citet{wen2014sentiment} use sentiment analysis to monitor students' trending opinions towards the course and to correlate sentiment with dropouts over time using survival analysis.
\citet{yang2015exploring} predict students' confusion with learning activities as expressed in the discussion forums using discussion behaviour and clickstream data, and explore the impact of confusion on student dropout. 
\citet{ramesh2015weakly} predict sentiment in MOOC forums using hinge-loss Markov random fields. \citet{yang2014question} study question recommendation in discussion forums based on matrix factorisation. 
\citet{gillani2014communication} find communities using Bayesian Non-Negative Matrix Factorisation. Despite this variety of works, no machine learning research has explored forum discussions for the purpose of measurement in MOOCs.

Topic modelling has been applied in MOOCs for tasks such as understanding key themes in forum discussions~\cite{robinson2015exploring}, predicting student survival~\cite{ramesh2014understanding}, study partner recommendation~\cite{xu2015study} and course recommendation~\cite{apazaonline}.
However, to our knowledge, no studies have leveraged topic modelling for measurement. More generally, psychometric models have enjoyed only fleeting attention by the machine learning community previously.

\section{Preliminaries and Problem Formalisation}
We choose NMF as the basic approach to discover forum topics due to the interpretability of topics produced, and the extensibility of its optimisation program. We begin with a brief overview of NMF and then define our problem.
\subsection{Non-Negative Matrix Factorisation (NMF)}
Given a non-negative matrix $\vbf{V}\in \mathbb{R}^{m\times n}$  and a positive integer $k$, NMF factorises $\vbf{V}$ into the product of a non-negative matrix $\vbf{W}\in \mathbb{R}^{m\times k}$ and a non-negative matrix $\vbf{H}\in \mathbb{R}^{k\times n}$
\begin{eqnarray*}
    \vbf{V}\approx \vbf{WH}
\end{eqnarray*}
A commonly-used measure for quantifying the quality of this approximation is the Frobenius norm between $\vbf{V}$ and $\vbf{WH}$. Thus, NMF involves solving the following optimisation problem,
\begin{eqnarray}
\min_{\vbf{W},\vbf{H}} 
%J(\vbf{W},\vbf{H})
%\mathcal{O}
\norm{\vbf{V}-\vbf{WH}}  
&\st& \vbf{W}\geq \vbf{0},\ \ \vbf{H}\geq \vbf{0} \enspace. \label{equ:nmf}
\end{eqnarray}
The objective function is convex in $\vbf{W}$ and $\vbf{H}$ separately, but not together. Therefore standard optimisers are not expected to find a global optimum. The multiplicative update algorithm~\cite{{lee2001algorithms}} is commonly used to find a local optimum,
where $\vbf{W}$ and $\vbf{H}$ are updated by a multiplicative factor that depends on the quality of the approximation. 

\subsection{Problem Statement}
We explore the automatic discovery of forum discussion topics for measurement in MOOCs. 
Our central tenet is that topics can be regarded as useful items for measuring a latent skill, if student responses to these items conform to a Guttman scale, and if the topics are semantically-meaningful to domain experts. 
As Guttman scale item responses are typically dichotomous, we consider item responses to be whether a student posts on the topic or not. 
Our goal is to generate a set of meaningful topics
that yield a student-topic matrix conforming to the properties of a \textit{Guttman scale}, 
\eg a near-triangular matrix (see Table~\ref{tab:guttman}). This process can be cast as optimisation.
We apply such well-scaled topics to measure skill attainment---as level of forum participation is known to be predictive of learning outcomes~\cite{Beaudoin2002147}.

Using NMF, a word-student matrix $\vbf{V}$ can be factorised into two   non-negative matrices: word-topic matrix $\vbf{W}$ and topic-student matrix $\vbf{H}$.
Our application requires that the topic-student matrix $\vbf{H}$ be \textbf{a) Binary} ensuring the response of a student to a topic is dichotomous; and \textbf{b) Guttman-scaled} ensuring the student responses to topics conform to a Guttman scale. NMF provides an elegant framework for incorporating these educational constraints via adding novel regularisation, as detailed in the next section.
A glossary of important symbols used in this paper is given in Table~\ref{tab:symbols}.
\begin{table}[ht]
    \small
    \centering
    \caption{Glossary of symbols}
    \label{tab:symbols}
    \begin{tabular}{lll}
        \toprule
        Symbol		& 	Description \\
        \toprule
        $m$							    & the number of words \\ 
        $n$							    & the number of students \\
        $k$							    & the number of topics \\
        $\vbf{V}=(v_{ij})_{m\times n}$	&	word-student matrix \\
        $\vbf{W}=(w_{ij})_{m\times k}$	&	word-topic matrix \\
        $\vbf{H}=(h_{ij})_{k\times n}$	&	topic-student matrix \\
        $\vbf{H}_{ideal}=((h_{ideal})_{ij})_{k\times n}$	&	exemplar topic-student matrix \\
        & with ideal Guttman scale \\
        $\lambda_0, \lambda_1, \lambda_2$							    & regularisation coefficients \\
        \bottomrule	
    \end{tabular}
\end{table}

\section{NMF for Guttman scale (NMF-Guttman)}
\subsection{Primal Program}
We introduce the following regularisation terms on $\vbf{W}$ to prevent overfitting, and on $\vbf{H}$ to encourage a binary solution and Guttman scaling:
\begin{itemize}
    \item $\norm{\vbf{W}}$ to prevent overfitting;
    \item $\norm{\vbf{H}-\vbf{H}_{ideal}}$ to encourage a Guttman-scaled $\vbf{H}$, where $\vbf{H}_{ideal}$ is a constant matrix with ideal Guttman scale;
    \item $\norm{\vbf{H}\circ\vbf{H}-\vbf{H}}$ to encourage a binary solution $\vbf{H}$, where operator $\circ$ denotes the Hadamard product.
\end{itemize}
Binary matrix factorisation (BMF) is a variation of NMF, where the input matrix and the two factorised matrices are all binary. 
Inspired by the approach of \citet{zhang2007binary} and \citet{zhang2010binary}, we add regularisation term $\norm{\vbf{H}\circ\vbf{H}-\vbf{H}}$. Noting this term equals $\norm{\vbf{H}\circ\left(\vbf{H}-\vbf{1}\right)}$, it is clearly minimised by binary $\vbf{H}$.

These terms together yield the objective function 
\begin{equation}\label{equ:nmfguttmanobj}
\begin{aligned}
f(\vbf{W},\vbf{H}) =& \norm{\vbf{V}-\vbf{WH}} + 
\lambda_0\norm{\vbf{W}} \\
& +
\lambda_1\norm{\vbf{H}-\vbf{H}_{ideal}} +
\lambda_2\norm{\vbf{H}\circ\vbf{H}-\vbf{H}}\enspace,
\end{aligned}\end{equation}
where $\lambda_0, \lambda_1, \lambda_2>0$ are regularisation parameters; with primal program
\begin{equation}\label{equ:nmfguttman}
\min_{\vbf{W},\vbf{H}} f(\vbf{W},\vbf{H}) 
\ \ \  \mbox{s.t.} \ \ \   \vbf{W}\geq \vbf{0},\ \vbf{H}\geq \vbf{0}\enspace.
\end{equation}

\subsection{Algorithm}
A local optimum of program~\eqref{equ:nmfguttman} is achieved via iteration
\begin{eqnarray}
\hspace{-1.0em}	w_{ij} \hspace{-1.0em}&\leftarrow&
\hspace{-1.0em}w_{ij}
\frac{(\vbf{V}\vbf{H}^T)_{ij}}
{(\vbf{WH}\vbf{H}^T+\lambda_0\vbf{W})_{ij}} \label{equ:w} \\
%H
\hspace{-1.0em}h_{ij} \hspace{-1.0em}&\leftarrow&
\hspace{-1.0em}h_{ij}
\frac{(\vbf{W}^T\vbf{V})_{ij}+4\lambda_2h_{ij}^3+3\lambda_2h_{ij}^2+\lambda_1(h_{ideal})_{ij}}
{(\vbf{W}^T\vbf{WH})_{ij}+6\lambda_2h_{ij}^3+(\lambda_1 + \lambda_2)h_{ij}} \label{equ:h}
\end{eqnarray}

These rules for the constrained program can be derived via the Karush-Kuhn-Tucker conditions necessary for
local optimality. First we construct the unconstrained Lagrangian 
\begin{equation*}
\mathcal{L}(\vbf{W},\vbf{H},\vbf{\alpha},\vbf{\beta})=f(\vbf{W},\vbf{H})+\text{tr}(\vbf{\alpha}\vbf{W})+\text{tr}(\vbf{\beta}\vbf{H})\enspace,
\end{equation*}
where $\alpha_{ij}, \beta_{ij}\leq 0$ are the Lagrangian dual variables for inequality constraints $w_{ij}\geq 0$ and $h_{ij}\geq 0$ respectively, and $\vbf{\alpha}=[\alpha_{ij}]$, $\vbf{\beta}=[\beta_{ij}]$ denote their corresponding matrices.

The KKT condition of stationarity requires  that the derivative of $\mathcal{L}$ with respect to $\vbf{W},\vbf{H}$ vanishes:
\begin{equation*}
\begin{split}
\frac{\partial \mathcal{L}}{\partial \vbf{W}}=&2\left(\vbf{W}^\star\vbf{H}^\star{\vbf{H}}^{\star T}- \vbf{V}\vbf{H}^{\star T} + \lambda_0\vbf{W}^\star\right)  + \vbf{\alpha}^\star = \vbf{0} \enspace, \\
\frac{\partial \mathcal{L}}{\partial \vbf{H}} = & 2\Big(
\vbf{W}^{\star T}\vbf{W}^\star\vbf{H}^\star 
- \vbf{W}^{\star T}\vbf{V}
+ (\lambda_1+\lambda_2) \vbf{H}^\star  \\
 &- \lambda_1 \vbf{H}_{ideal} \Big)
+ 4\lambda_{2}\vbf{H}^\star\circ\vbf{H}^\star\circ\vbf{H}^\star\\
&- 6\lambda_2\vbf{H}^\star\circ\vbf{H}^\star 
+ \vbf{\beta}^\star
= \vbf{0}\enspace.
\end{split}\end{equation*}
Complementary slackness $\alpha^\star_{ij}w^\star_{ij}=\beta^\star_{ij}h^\star_{ij}=0$, implies:
\begin{eqnarray*}
    0 \hspace{-1em}&=& \hspace{-1em}\left(\vbf{V}\vbf{H}^{\star T} - \vbf{W}^\star\vbf{H}^\star\vbf{H}^{\star T} - \lambda_0\vbf{W}^\star\right)_{ij}w^\star_{ij} \enspace, \label{equ:W} \\
    0 \hspace{-1em}&= & \hspace{-1em} \Big(\vbf{W}^{\star T}\vbf{V} +
    3\lambda_2\vbf{H}^\star\circ\vbf{H}^\star + \lambda_1 \vbf{H}_{ideal} -\vbf{W}^{\star T}\vbf{W}^\star\vbf{H}^\star %\right.  
    \nonumber \\
    \hspace{-1em} && \hspace{-1em} %\left.   -
    -2\lambda_{2}\vbf{H}^\star\circ\vbf{H}^\star\circ\vbf{H}^\star 
    - (\lambda_1+\lambda_2) \vbf{H}^\star %\right.
    \nonumber\\
    \hspace{-1em} && \hspace{-1em} %\left.
    + 4\lambda_{2}\vbf{H}^\star\circ\vbf{H}^\star\circ\vbf{H}^\star
    -
    4\lambda_{2}\vbf{H}^\star\circ\vbf{H}^\star\circ\vbf{H}^\star
    \Big)_{ij}h^\star_{ij} \enspace. \label{equ:H}
\end{eqnarray*}
These two equations lead to the updating rules~\eqref{equ:w}, \eqref{equ:h}. 
Our next result proves that these rules improve the objective value.
\begin{thm}\label{thm:thm1}
    The objective function $f(\vbf{W},\vbf{H})$ of program~\eqref{equ:nmfguttman} is non-increasing under update rules~\eqref{equ:w} and~\eqref{equ:h}.
\end{thm}

The proof of Theorem~\ref{thm:thm1} is given in
\iftoggle{fullpaper}{the Appendix.}{\citet{he2016moocs}.}
Our overall approach is described as Algorithm~\ref{alg:algnmf}.
$\vbf{W}$ and $\vbf{H}$ are initialised using plain NMF~\cite{lee1999learning,lee2001algorithms}, then normalised~\cite{zhang2007binary,zhang2010binary}.

\begin{algorithm}[h]
    \caption{NMF-Guttman} \label{alg:algnmf}
    \renewcommand{\arraystretch}{2}
    \begin{algorithmic}[1]
        \REQUIRE ~~\\
        $\vbf{V}$, $\vbf{H}_{ideal}$, $\lambda_0$, $\lambda_1$, $\lambda_2$, $k$;
        \ENSURE ~~\\
        A topic-student matrix, $\vbf{H}$;
        
        \STATE Initialise $\vbf{W}$, $\vbf{H}$ using NMF;
        \STATE Normalise $\vbf{W}$, $\vbf{H}$ following ~\cite{zhang2007binary,zhang2010binary};
        \REPEAT
        \STATE Update $\vbf{W},\vbf{H}$ iteratively based on Eq.~(\ref{equ:w}) and Eq.~(\ref{equ:h});
        \UNTIL{converged}
        \RETURN $\vbf{H}$;
    \end{algorithmic}
\end{algorithm}

\subsubsection{Selection of $\vbf{H}_{ideal}$}
Topic-student matrix $\vbf{H}_{ideal}$ is an ideal target where students' topic responses conform to a perfect Guttman scale. 
$\vbf{H}_{ideal}$ can be obtained in different ways depending on the attribute of interest to be measured.
In this paper, we are interested in measuring students' latent skill in MOOCs. We envision measurement at the completion of a first offering, with scaled items applied in subsequent offerings for measuring students or curriculum design; alternatively within one offering after a mid-term.
Thus, $\vbf{H}_{ideal}$ can be obtained using assessment, \emph{which need not be based on Guttman-scaled items}.
For each student $j$, his/her responses to the topics given by column $(h_{ideal})_{\cdot j}$ are selected based on his/her grade $g_j\in[0,100]$, as
\begin{eqnarray*}
    &&(h_{ideal})_{\cdot j}= (
    \rlap{$\underbrace{1\cdots 1}_{b}$}\;\;\;\;\;\;\;\;\;\;
    \rlap{$\underbrace{0\cdots 0}_{k-b}$}\;\;\;\;\;\;\;\;\;) \\ %\label{equ:inihideal}
    \mbox{where} && b=\min\left\{\floor*{\frac{g_j+width}{width}}, k\right\},
    width=\frac{100}{k} \enspace. \nonumber
\end{eqnarray*}
For example, student $j$ with $g_j=35$ has response pattern on $k=10$ topics $(h_{ideal})_{.j}=(1111000000)$. 

\section{Experiments}
We conduct experiments to evaluate the effectiveness of our algorithms on real MOOCs on Coursera. We also demonstrate the robustness of our approach in terms of parameter sensitivity.
In our experiments, we use the first offerings of three Coursera MOOCs from Education, Economics and Computer Science offered by The University of Melbourne. They are \textit{Assessment and Teaching of 21st Century Skills}, \textit{Principles of Macroeconomics}, \textit{Discrete Optimisation} and are named EDU, ECON and OPT for short respectively.

\subsection{Dataset Preparation}
We focus on the students who contributed posts or comments in forums.
For each student, we aggregate all posts/comments that s/he contributed.
After stemming, removing stop words and html tags, a word-student matrix with normalised tf-idf in [0,1] is produced.
The statistics of words and students for the MOOCs are displayed in Table~\ref{tab:stat}.
\begin{table}[ht!]
    \centering
    \caption{Statistics of Datasets}
    \label{tab:stat}
    \begin{tabular}{lcc} 
        \toprule
        MOOC  & \#Words & \#Students\\
        \toprule
        EDU & 20,126 & 1,749\\ 
        ECON & 22,707& 1,551\\ 
        OPT & 17,059  & 1,092\\ 
        \bottomrule\end{tabular}
\end{table}

\subsection{Baseline Approach and Evaluation Metrics}
Since there has been no prior method to automatically generate topics forming a Guttman scale, we compare our algorithm with standard NMF (with no regularisation on $\vbf{H}_{ideal}$).

We adopt the Coefficient of Reproducibility (CR) as it is commonly used to evaluate Guttman scale quality:
\begin{eqnarray*}
    %\small
    CR=1-\frac{\text{No. of errors}}{\text{No. of possible errors(Total responses)}}\enspace.
\end{eqnarray*}
CR measures how well a student's responses can be predicted given his/her position on the scale, \ie total score. By convention, a scale is accepted with items scaled unidimensionally, if its CR is at least 0.90~\cite{guttman1950basis}.

To guarantee binary $\vbf{H}$, we first scale to $\frac{h_{ij}-\min(\vbf{H})}{\max(\vbf{H})-\min(\vbf{H})}\in[0,1]$, then threshold against a value in $[0.1, 0.2, \cdots, 0.9]$ maximising  CR, so that we \emph{conservatively} report CR.

\subsection{Experimental Setup}
\subsubsection{Evaluation Setting}
We split data into a training set (70\% students) and a test set (30\% students). The topics are generated by optimising the objective function \eqref{equ:nmfguttmanobj} on the training set, and evaluated using CR and the quality of approximation $\norm{\vbf{V}-\vbf{WH}}$. 
To simulate the inferring responses for new students, 
which has not been explored previously, 
the trained model is evaluated on the test set using Precision-Recall and ROC curves. Note that in the psychometric literature, validation typically ends with an accepted ($>0.9$) CR on the training set.

After learning on the training set word-student matrix $\vbf{V}^{(train)}$, two matrices are produced: a word-topic matrix $\vbf{W}^{(train)}$ and topic-student matrix $\vbf{H}^{(train)}$.
To evaluate 
the trained model
on the test set $\vbf{V}^{(test)}$, we apply the trained word-topic matrix $\vbf{W}^{(train)}$. Together, we have the relations
\begin{eqnarray*}
    \vbf{V}^{(train)} &=& \vbf{W}^{*(train)}\vbf{H}^{(train)}\\
    \vbf{V}^{(test)} &=& \vbf{W}^{*(train)}\vbf{H}^{(test)}\enspace.
\end{eqnarray*}
Solving for $\vbf{H}^{(test)}$ yields
\begin{eqnarray*}
    \vbf{H}^{(test)}=\vbf{H}^{(train)}(\vbf{V}^{(train)})^{\dagger}\vbf{V}^{(test)}\enspace.
\end{eqnarray*}
where $(\vbf{V}^{(train)})^{\dagger}$ denotes the pseudoinverse of $\vbf{V}^{(train)}$.
\subsubsection{Hyperparameter Settings}
Table~\ref{tab:paras} shows the parameter values used for parameter sensitivity experiments, where the default values in boldface are used in other experiments.
\begin{table}[ht]
    \centering
    \caption{Hyperparameter Settings}
    \label{tab:paras}
    \begin{tabular}{lll}
        \toprule
        Parameter		& 	Values Explored (\textbf{Default Value}) \\
        \toprule
        $\lambda_0$	&	$[10^{-4},10^{-3},10^{-2},\boldsymbol{10^{-1}},10^{0},10^{1},10^{2}]$ \\
        $\lambda_1$	&	$[10^{-4},10^{-3},10^{-2},\boldsymbol{10^{-1}},10^{0},10^{1},10^{2}]$ \\
        $\lambda_2$	&	$[10^{-4},10^{-3},\boldsymbol{10^{-2}},10^{-1},10^{0},10^{1},10^{2}]$ \\
        $k$	&	$[5,\boldsymbol{10},15,20,25,30]$ \\
        \bottomrule
    \end{tabular}
\end{table}

\subsection{Results}
\begin{figure}[!htb]
    \centering
    \begin{subfigure}[t]{0.23\textwidth}
        \centering
        \includegraphics[scale=0.275]{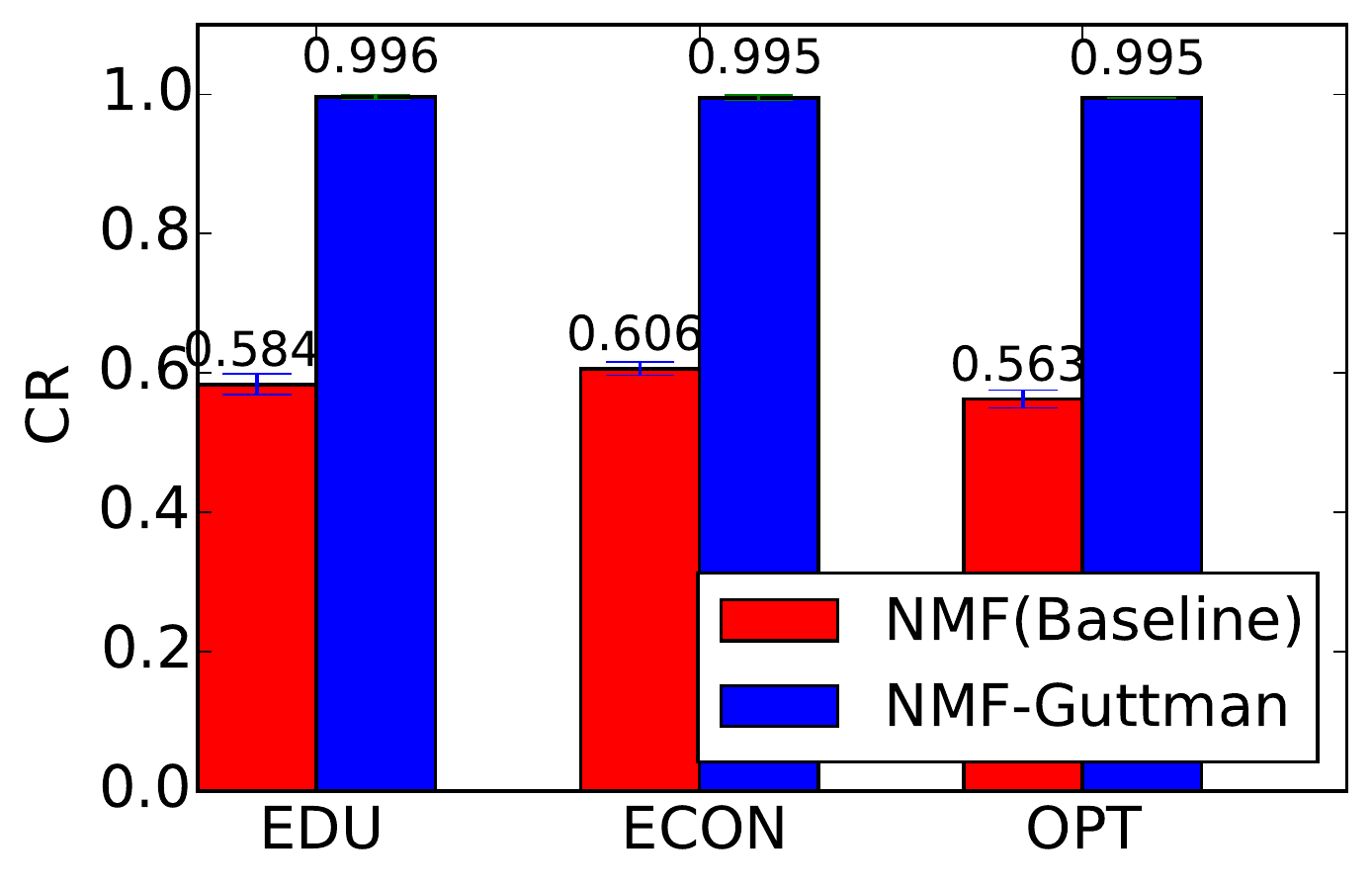}
        \caption{CR}
    \end{subfigure}%
    ~ 
    \begin{subfigure}[t]{0.23\textwidth}
        \centering
        \includegraphics[scale=0.275]{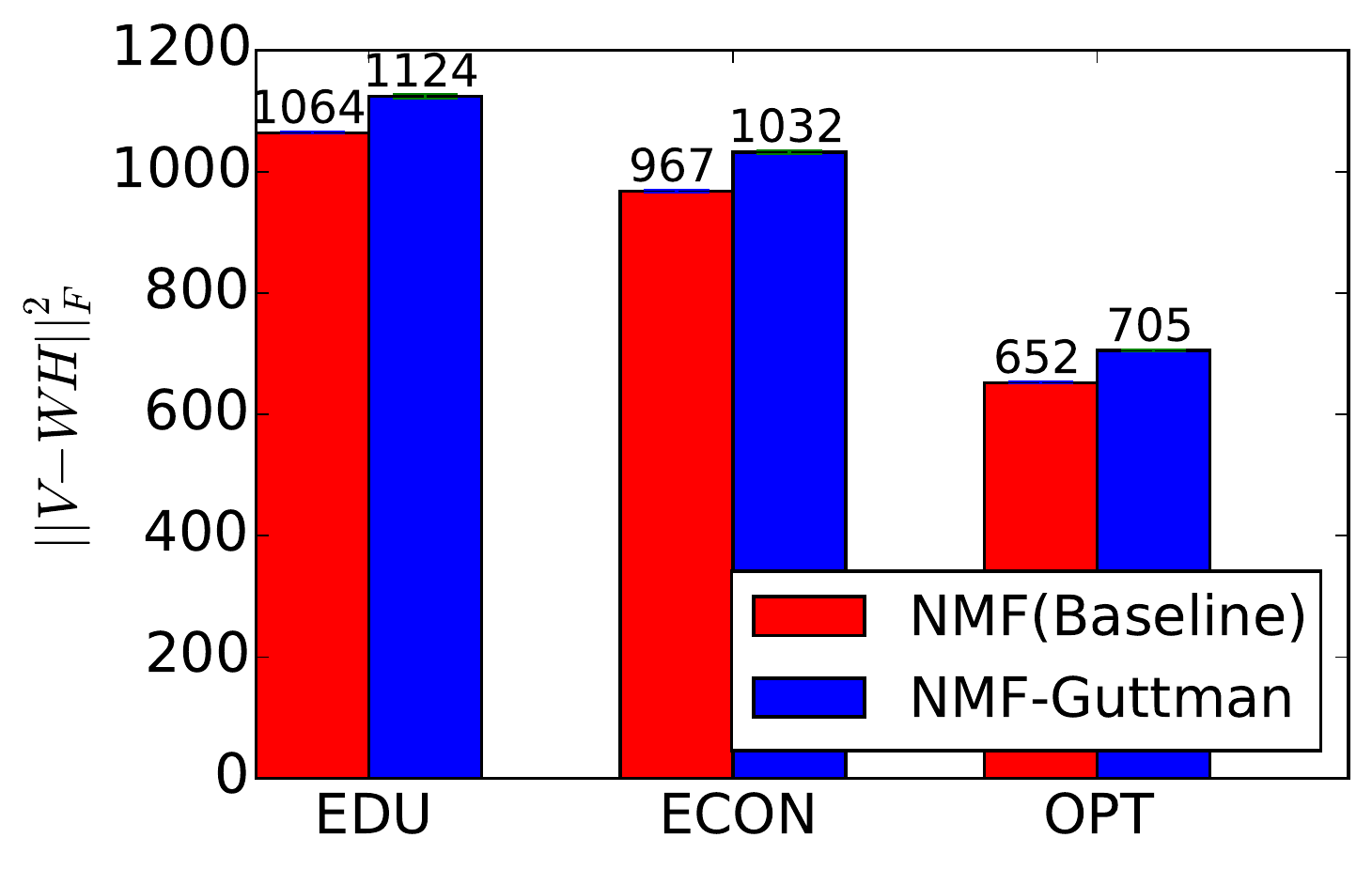}
        \caption{$\norm{\vbf{V}-\vbf{WH}}$}
    \end{subfigure}
    \caption{Comparison of NMF and NMF-Guttman in terms of CR and $\norm{\vbf{V}-\vbf{WH}}$.}
    \label{fig:results}
\end{figure}

In this group of experiments, we examine how well the generated topics conform to a Guttman scale, and the quality of approximation $\vbf{WH}$ to $\vbf{V}$.
The reported results are the results averaged over 10 runs. The parameters are set using the values in boldface in Table~\ref{tab:paras}. Figure~\ref{fig:results} displays the comparison between our algorithm NMF-Guttman and the baseline NMF in terms of CR, and the quality of approximation $\vbf{WH}$ to $\vbf{V}$ on the training set. 

It is clear that our algorithm NMF-Guttman can provide excellent performance in terms of CR with nearly a perfect 1.0, well above the 0.9 cutoff for acceptance. This significantly outperforms baseline which has ~0.60 CR across the MOOCs, below Guttman scale acceptance. Meanwhile, NMF-Guttman maintains good quality of approximation, with only slightly inferior $\norm{\vbf{V}-\vbf{WH}}$ comparing to NMF (5\%, 6\%, 8\% worse on EDU, ECON, OPT).
This is reasonable, as NMF-Guttman has more constraints hence the model itself is less likely to approximate $\vbf{V}$ as well as the less constrained standard NMF.

\begin{figure}[!htb]
    \centering
    \begin{subfigure}[t]{0.23\textwidth}
        \centering
        \includegraphics[scale=0.418]{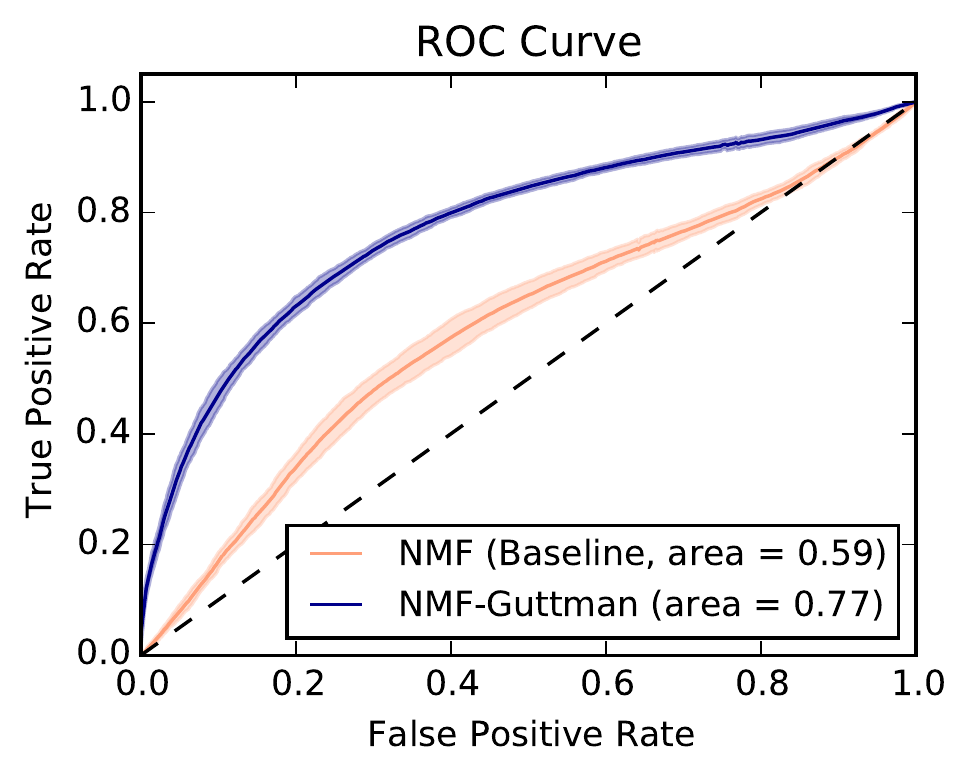}
        \caption{ECON}
    \end{subfigure}%
    ~ 
    \begin{subfigure}[t]{0.23\textwidth}
        \centering
        \includegraphics[scale=0.418]{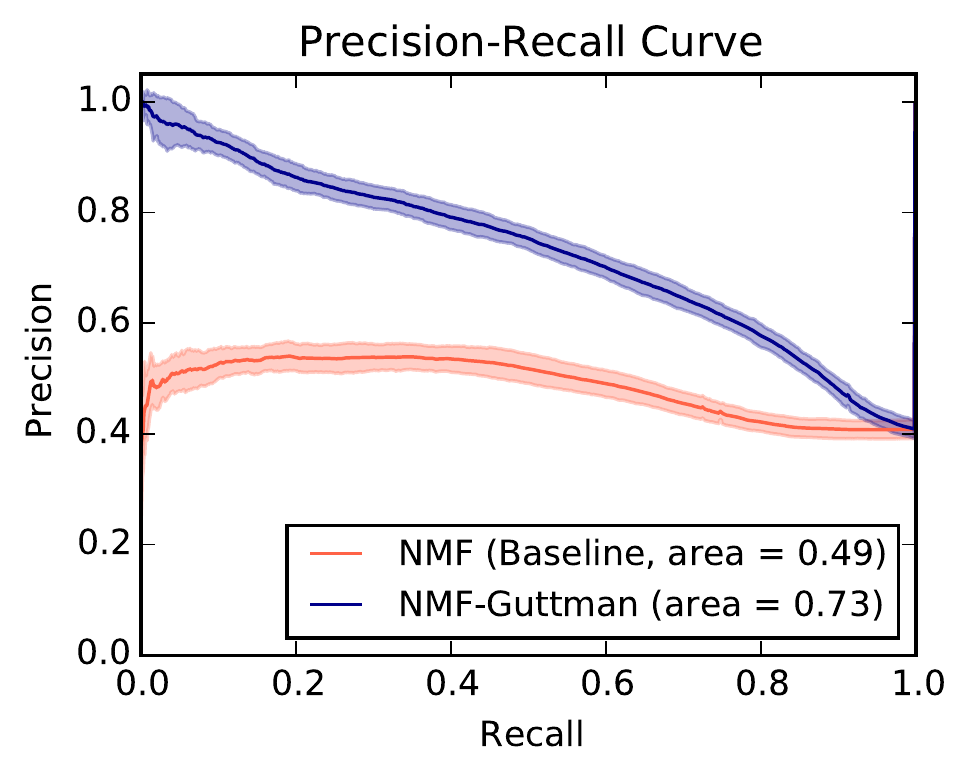}
        \caption{ECON}
    \end{subfigure}
    \caption{Comparison of NMF and NMF-Guttman in terms of ROC curve and Precision-Recall curve.}
    \label{fig:rocs}
\end{figure}

The ROC and Precision-Recall curves (averaged curves with standard deviation over 10 runs) on test set for the ECON MOOC are shown in Figure~\ref{fig:rocs}.
It can be seen that NMF-Guttman significantly dominates NMF, with around 20\%-30\% better performance, demonstrating the possibility of using the topics for inferring the response of unseen students. Similar results can be found on the remaining MOOCs in 
\iftoggle{fullpaper}{the Appendix.}{\citet{he2016moocs}.}

\begin{figure}[!htb]
    \centering
    \begin{subfigure}[t]{0.23\textwidth}
        \centering
        \includegraphics[scale=0.337]{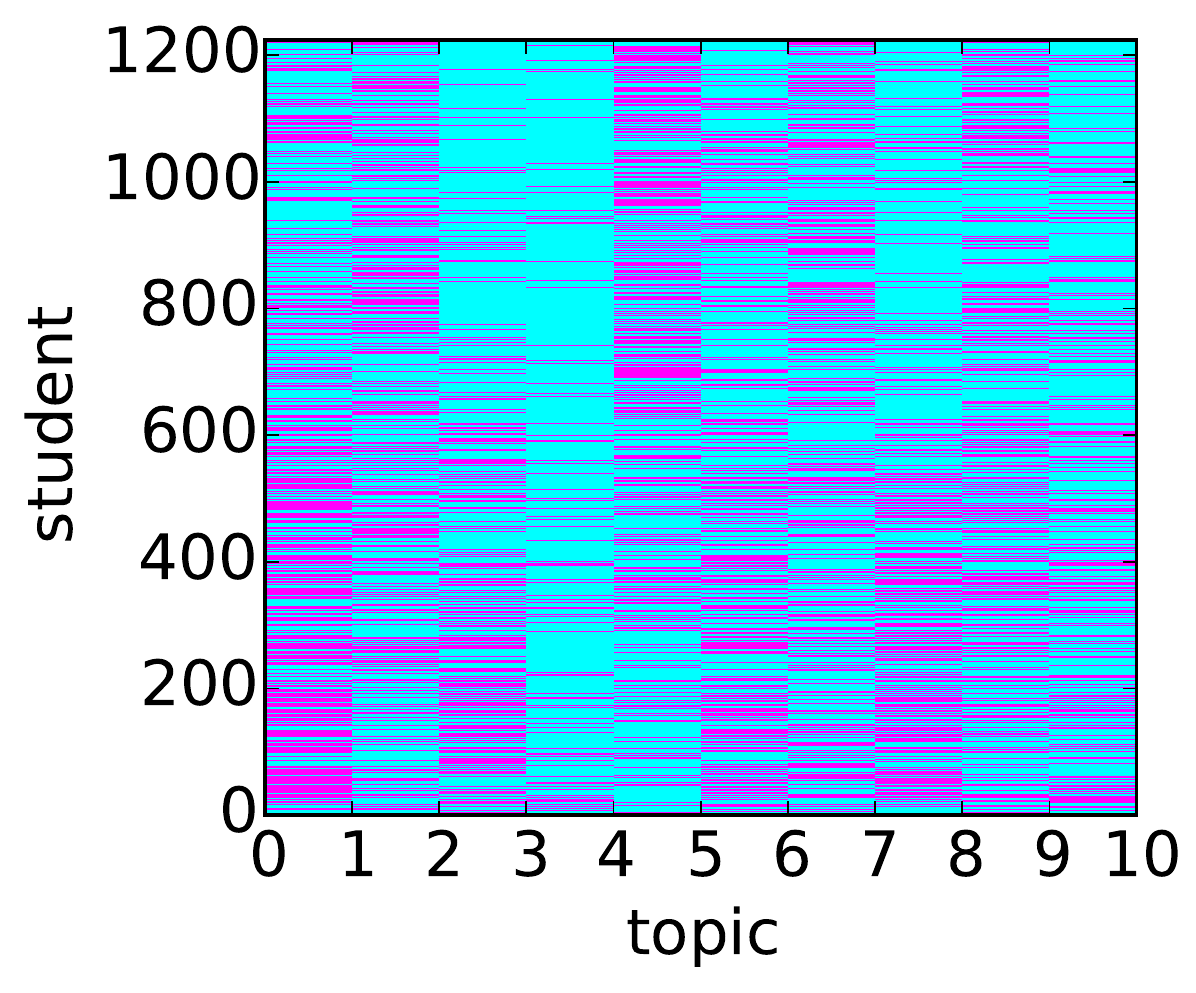}
        \caption{NMF}
    \end{subfigure}%
    ~ 
    \begin{subfigure}[t]{0.23\textwidth}
        \centering
        \includegraphics[scale=0.337]{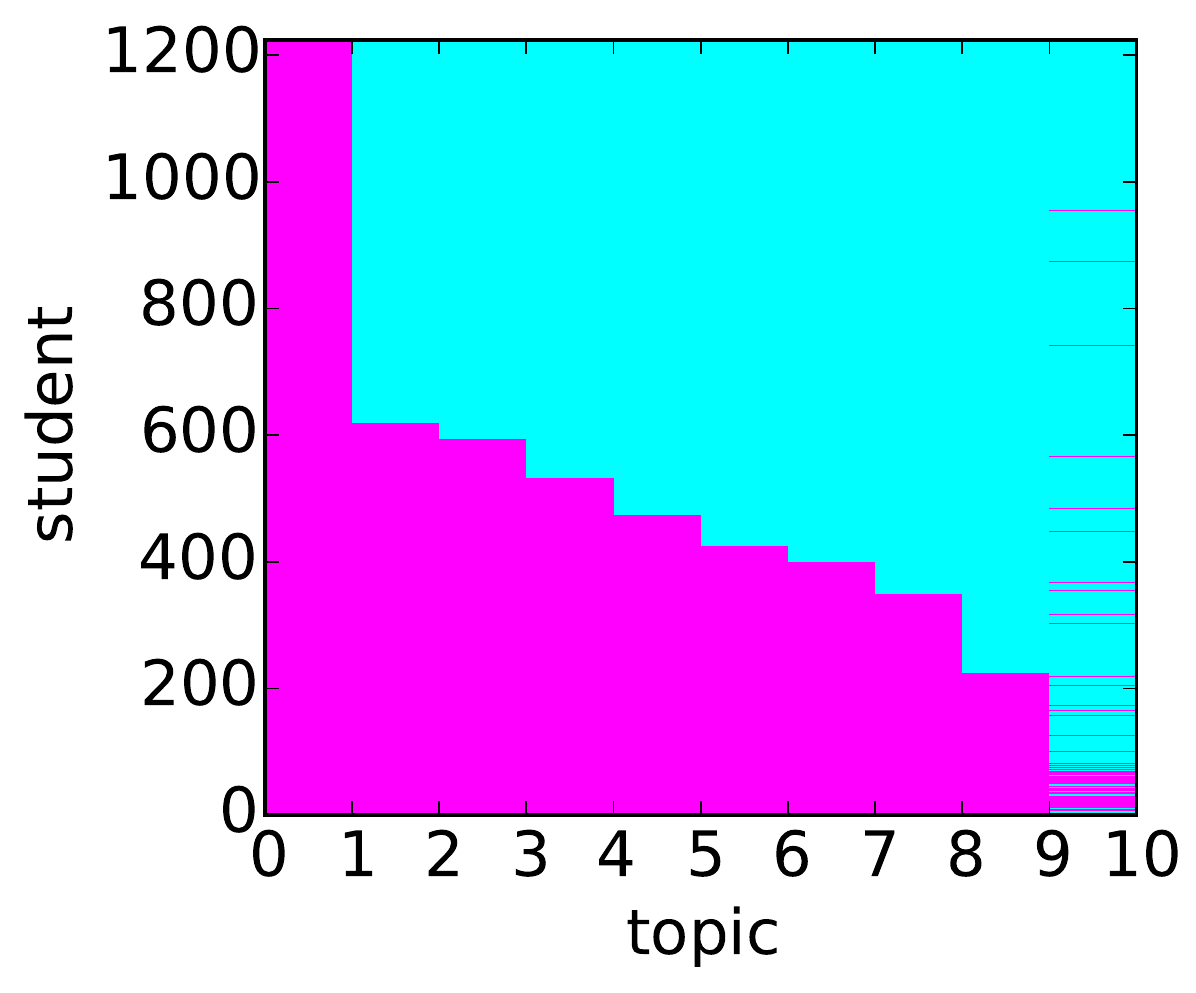}
        \caption{NMF-Guttman}
        \label{fig:guttmatrix}
    \end{subfigure}
    \caption{Student-topic matrix generated by NMF and NMF-Guttman for MOOC EDU; fuchsia for 1, cyan for 0.}
    \label{fig:vistrainbio}
\end{figure}

We next visualise the student-topic matrix $\vbf{H}^T$ produced by NMF and NMF-Guttman respectively. Figure~\ref{fig:vistrainbio} is a clear demonstration that NMF-Guttman can produce excellent Guttman scales, while NMF may not. 
Around half of the cohort (having grade=0) only contribute to topic 1---the easiest---while only a few students contribute to topic 10---the most difficult. 

NMF-Guttman can discover items (topics) with responses conforming to a Guttman scale while maintaining the quality of factorisation approximation. 
It also effectively infers new students' responses. 

\subsection{Validity} 
The results above establish that our algorithm generates items (topics) with responses conforming to the Guttman scale. Next we test validity---whether topics are meaningful in two aspects: \textbf{a) Interpretability}: Are the topics interpretable? \textbf{b) Difficulty level}: Do topics exhibit different difficulty levels as inferred by our algorithm and implied by the Guttman scale?

\subsubsection{Qualitative Survey}
To answer the above questions, we interviewed experts with relevant course background. We showed them the topics (each topic is represented by top 10 words) discovered by our algorithm NMF-Guttman and those generated from the baseline NMF, while blinding interviewees to the topic set's source. 
We randomised topic orders, and algorithm order, since our algorithm naturally suggests topic order. We posed the following questions for the topics from NMF-Guttman and NMF respectively:
\begin{enumerate}[Q1.]
    \item \emph{Interpretation:} interpret the topic's meaning based on its top 10 word description.
    \item \emph{Interpretability:} how easy is interpretation? 1=Very difficult; 2=Difficult; 3=Neutral; 4=Easy; 5=Very easy.
    \item \emph{Difficulty level:} how difficult is the topic to learn? 1=Very easy; 2=Easy; 3=Neutral; 4=Difficult; 5=Very difficult.
    \item \emph{Ranking:} rank the topics according to their difficulty levels. From 1=easiest; to 10=most difficult.
\end{enumerate}

\textbf{a) OPT MOOC}
We interviewed 5 experts with PhDs in discrete optimisation for the OPT MOOC. 
To validate the topics' difficulty levels, we compute the Spearman's rank correlation coefficient between the ranking from our algorithm and the one from each interviewee, which is shown in Table~\ref{tab:survey}. 
There is high correlation between the NMF-Guttman ranking and those of the Interviewees, \emph{suggesting the topics' Guttman scale relates to difficulty.}
\begin{table}[!htb]
    \small
    \centering
    \caption{Survey for OPT MOOC.}
    \begin{tabular}{clcc}
        \toprule
        \multirow{2}{*}{Interviewee}  & \multirow{2}{*}{Background} & Spearman's rank  \\
        %\cline{3-4}
        %& 
        & & correlation coefficient\\
        \toprule
        \multirow{2}{*}{1}   & Works in optimisation &  \multirow{2}{*}{0.71} \\
        & and took OPT MOOC & \\
        %\hline
        2   & Tutor for OPT MOOC & 0.37 \\
        %\hline
        \multirow{2}{*}{3}    & Professor who teaches &\multirow{2}{*}{0.90} \\
        & optimisation courses &  \\
        %\hline
        4 & Works in optimisation   & 0.67  \\
        %\hline
        5  & Works in optimisation  & 0.41 \\
        \bottomrule
    \end{tabular}%
    \label{tab:survey}%
\end{table}%

Table~\ref{tab:interpretation} depicts the interpretation on a selection of four topics by Interviewee 1, who took the OPT MOOC previously.
The compelete interpretation for the topics from NMF and NMF-Guttman can be found in
\iftoggle{fullpaper}{the Appendix.}{\citet{he2016moocs}.}
\begin{table*}
    \small
    \centering
    \caption{Interviewee 1's interpretation on OPT MOOC topics generated from NMF-Guttman with inferred difficulty ranking.}
    \begin{tabular}{clll}
        \toprule
        No. &  Topics  & Interpretation & Inferred Ranking \\
        \toprule
        1 & python problem file solver assign pi class video course use & How to use platform/python & 1 (Easiest)  \\
        2 &  submit thank please pyc grade feedback run solution check object & Platform/submission issues & 2 \\
        5 &  color opt random search local greedy swap node good get & Understand and implement local search & 5\\
        8 & time temperature sa move opt would like well start ls & How to design and tune simulated & 8 \\
        &  &  \hspace{2em} annealing and local search \\
        \bottomrule
    \end{tabular}%
    \label{tab:interpretation}%
\end{table*}%
It can be seen that the topics from NMF-Guttman are interpretable and exhibit different difficulty levels, qualitatively validating the topics can be used to measure students' latent skill. 
Note that the topics produced by NMF do not conform to a Guttman scale and are not designed for measurement. Indeed we observed informally that NMF-Guttman's topics were more diverse than those of NMF. 
For OPT MOOC, half of the topics are not relevant to the course content directly, \ie feedback about the course and platform/submission issues.
While most of the topics from NMF-Guttman are closely relevant to the course content, which are more useful to measure students' skill or conduct curriculum refinement.

\textbf{b) EDU MOOC}
The course coordinator who has detailed understanding of the course, its curriculum and its forums, was interviewed
to answer our survey questions.
A 0.8 Spearman's rank correlation coefficient is found between the NMF-Guttman ranking and that of the course coordinator, supporting that the inferred difficulty levels are meaningful.
Furthermore, most of the NMF-Guttman's topics are interpretable, less fuzzy, and less overlapping than those of NMF, ad judged by the course coordinator.
The topic interpretations can be found in
\iftoggle{fullpaper}{the Appendix.}{\citet{he2016moocs}.}

\subsection{Parameter Sensitivity}\label{sec:para}
To validate the robustness of parameters and analyse the effect of the parameters, a group of experiments were conducted. The parameter settings are shown in Table~\ref{tab:paras}.
Due to space limitation, we only report the results for $\lambda_1$ on OPT MOOC. 
Results for parameter $\lambda_0$,$\lambda_1$,$\lambda_2$ and $k$ on all three MOOCs can be found in 
\iftoggle{fullpaper}{the Appendix.}{\citet{he2016moocs}.}

\subsubsection{Regularisation Parameter $\lambda_1$} 
\begin{figure}[!htb]
    \centering
    \begin{subfigure}[t]{0.23\textwidth}
        \centering
        \includegraphics[scale=0.31]{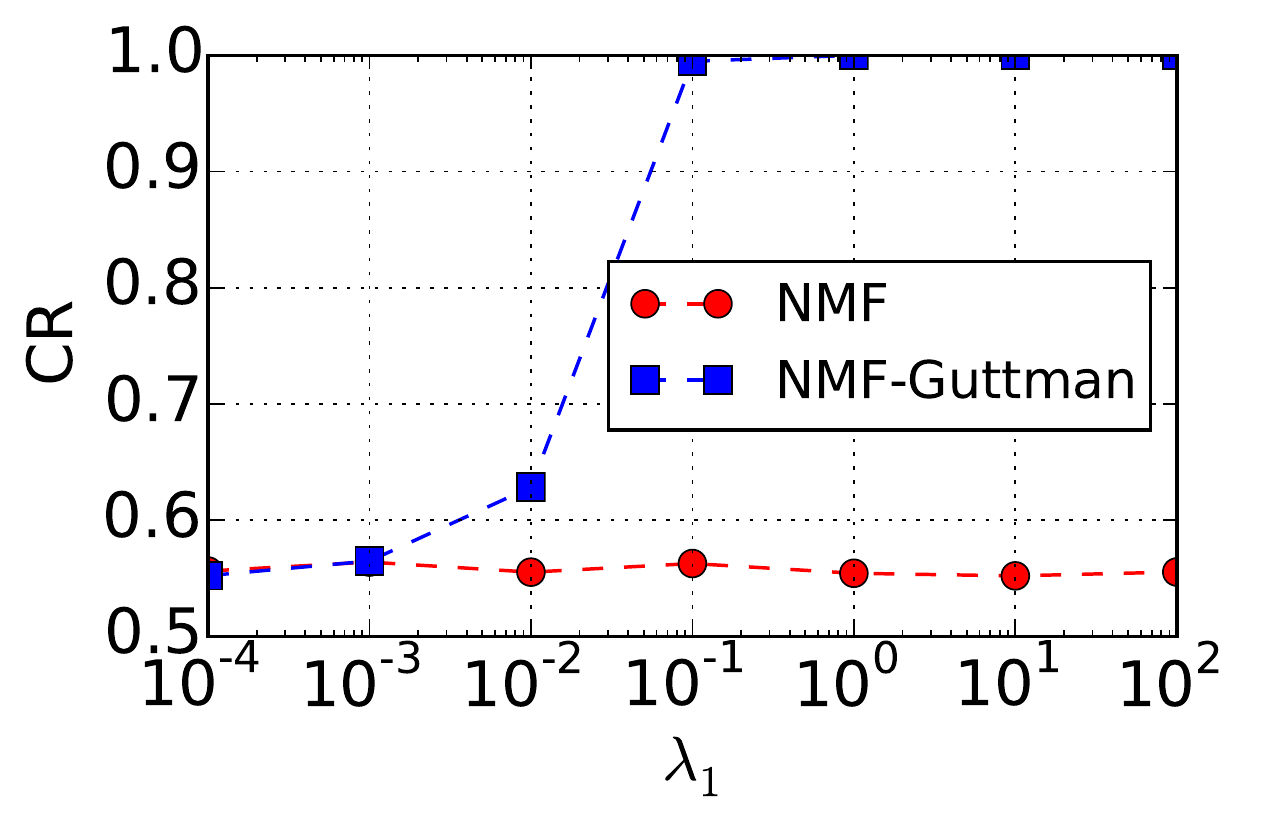}
        \caption{CR on OPT}
    \end{subfigure}%
    ~ 
    \begin{subfigure}[t]{0.23\textwidth}
        \centering
        \includegraphics[scale=0.31]{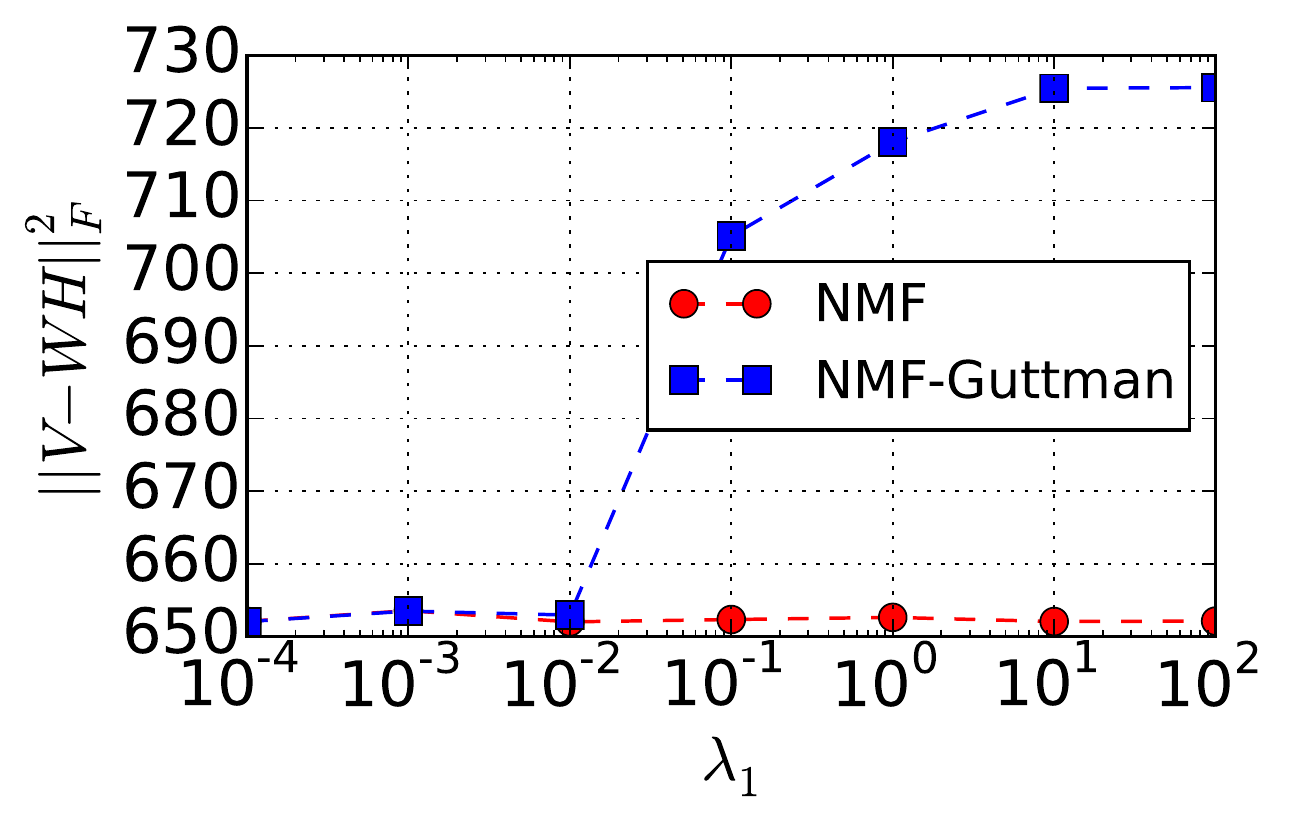}
        \caption{$\norm{\vbf{V-WH}}$ on OPT}
    \end{subfigure}
    \caption{Comparison of NMF and NMF-Guttman in terms of CR and $\norm{\vbf{V-WH}}$ with varying $\lambda_1$.}
    \label{fig:hideals}
\end{figure}

The performance of CR and $\norm{\vbf{V-WH}}$ with varying $\lambda_1$ is shown in Figure~\ref{fig:hideals}. It can be seen that NMF-Guttman's high performance is stable for $\lambda_1$ varying over a wide range $10^{-1}$ to $10^2$. 

Similar results are found for $\lambda_0$, $\lambda_2$, and $k$. NMF-Guttman is not sensitive to $\lambda_0$ and $k$. For $\lambda_2$, NMF-Guttman stably performs well when $\lambda_2$ varies from $10^{-4}$ to $10^{-2}$.

Overall, our algorithm NMF-Guttman is robust, consistently achieves much higher CR than NMF with varying $\lambda_0$, $\lambda_1$, $\lambda_2$ and  $k$, while maintaining the quality of approximation $\norm{\vbf{V-WH}}$.

\section{Conclusion}
This is the first study that combines a machine learning technique (topic modelling) with measurement theory (psychometrics) as used in education. Our focus is measurement for curriculum design and assessment in MOOCs. Motivated by findings that participation level in online forums is predictive of student performance~\cite{Beaudoin2002147}, we aim to automatically discover forum post topics on which student engagement forms a so-called Guttman scale: such scales are evidence of measuring educationally-meaningful skill attainment~\cite{guttman1950basis}. We achieve this goal by a novel regularisation of non-negative matrix factorisation.

Our empirical results are compelling and extensive. We contribute a quantitative validation on three Coursera MOOCs, demonstrating our algorithm conforms to Guttman scaling (shown with high coefficients of reproducibility), strong quality of factorisation approximation, and predictive power on unseen students (via ROC and PR curve analysis). We also contribute a qualitative study of domain expert interpretations on two MOOCs, showing that most of the topics with difficulty levels inferred, are interpretable and meaningful.

This paper opens a number of exciting directions for further research. Broadly speaking, the consequences of content-based measurement on educational theories and practice requires further understanding, while the study of statistical models for psychometrics by computer science will stimulate interesting new machine learning.

Our approach could be extended to incorporate partial prior knowledge. For example, an education researcher or instructor might already possess certain items for student engagement in MOOCs (\eg watching videos, clickstream observations, completing assignments, etc.) to measure some latent attribute. We are interested in exploring how to discover topics that measure the same attribute as measured by existing items. 

\bibliographystyle{aaai}
\bibliography{mandb}

\iftoggle{fullpaper}{
\section{Appendix}
\subsection{Proof of Theorem 1}
We follow the similar procedure described in \cite{lee2001algorithms}, where an auxiliary function similar to that used in the Expectation-Maximization (EM) algorithm is used for proof.
\begin{defn}
    \cite{lee2001algorithms} $G(h,h^\prime)$ is an auxiliary function for $F(h)$ if the conditions
    \begin{eqnarray*}
        G(h,h^\prime)\geq F(h), \qquad G(h,h)=F(h)
    \end{eqnarray*}
    are satisfied.
\end{defn}
\begin{lemma}
    \cite{lee2001algorithms} If $G$ is an auxiliary function, then $F$ is non-increasing under the update
    \begin{eqnarray}\label{equ:argmin}
        h_{t+1}=\argmin_hG(h,h^t) 
    \end{eqnarray}
    Proof: $F(h^{t+1})\leq G(h^{t+1}, h^t)\leq G(h^t, h^t)=F(h^t)$
\end{lemma}
For any element $h_{ij}$ in $\vbf{H}$, let $F_{h_{ij}}$ denote the part of $f(\vbf{W},\vbf{H})$ in Eq.~(\ref{equ:nmfguttmanobj}) in the paper relevant to $h_{ij}$. Since the update is essentially element-wise, it is sufficient to show that each $F_{h_{ij}}$ is non-increasing under the update rule of Eq.~(\ref{equ:h}) in the paper. To prove it, we define the auxiliary function regarding $h_{ij}$ as follows.
\begin{lemma}
    Function
    \begin{eqnarray}\label{equ:auxiliary}
        \begin{split}
            G(h_{ij}, h_{ij}^t)=& F_{h_{ij}}(h_{ij}^t)+F_{h_{ij}}^{\prime}(h_{ij}^t)(h_{ij}-h_{ij}^t)\\
            +&\varphi_{ij} (h_{ij}-h_{ij}^t)^2
        \end{split}
    \end{eqnarray}
    where
    \begin{eqnarray*}
        \varphi_{ij}=\frac{(\vbf{W}^T\vbf{WH})_{ij}+\lambda_1h_{ij}^t+6\lambda_2(h_{ij}^t)^3+\lambda_2h_{ij}^t}{h_{ij}^t}
    \end{eqnarray*} 
    is an auxiliary function for $F_{h_{ij}}$.
\end{lemma}
Proof: It is obvious that $G(h_{ij},h_{ij})=F_{h_{ij}}$. So we only need to prove that $G(h_{ij}, h_{ij}^t)\geq F_{h_{ij}}$. Considering the Taylor series expansion of $F_{h_{ij}}$,
\begin{eqnarray*}
    F_{h_{ij}}&=& F_{h_{ij}}(h_{ij}^t)+F_{h_{ij}}^{\prime}(h_{ij}^t)(h_{ij}-h_{ij}^t)\\
    &+&\frac{1}{2}F_{h_{ij}}^{\prime\prime}(h_{ij}^t)(h_{ij}-h_{ij}^t)^2 
\end{eqnarray*}
$G(h_{ij}, h_{ij}^t)\geq F_{h_{ij}}$ is equivalent to 
$\varphi_{ij}\geq \frac{1}{2}F_{h_{ij}}^{\prime\prime}(h_{ij}^t)$, where
\begin{eqnarray*}
    F_{h_{ij}}^{\prime\prime}(h_{ij}^t)&=&2(\vbf{W}^T\vbf{W})_{ii}+2\lambda_1+12\lambda_2(h_{ij}^t)^2\\
    &-&12\lambda_2h_{ij}^t+2\lambda_2
\end{eqnarray*}
To prove the above inequality, we have
\begin{eqnarray*}
    \varphi_{ij}h_{ij}^t\hspace{-1.0em}&=&\hspace{-1.0em}(\vbf{W}^T\vbf{WH})_{ij}+\lambda_1h_{ij}^t+6\lambda_2(h_{ij}^t)^3+\lambda_2h_{ij}^t\\
    \hspace{-1.0em}&=&\hspace{-1.0em}\sum_{l=1}^{k}(\vbf{W}^T\vbf{W})_{il}h_{lj}^t+\lambda_1h_{ij}^t+6\lambda_2(h_{ij}^t)^3+\lambda_2h_{ij}^t\\
    \hspace{-1.0em}&\geq&\hspace{-1.0em} (\vbf{W}^T\vbf{W})_{ii}h_{ij}^t+\lambda_1h_{ij}^t+6\lambda_2(h_{ij}^t)^3+\lambda_2h_{ij}^t\\	
    \hspace{-1.0em}&\geq&\hspace{-1.0em} h_{ij}^t\big((\vbf{W}^T\vbf{W})_{ii}+\lambda_1+6\lambda_2(h_{ij}^t)^2-6\lambda_2h_{ij}^t+\lambda_2\big)\\
    \hspace{-1.0em}&=&\hspace{-1.0em}\frac{1}{2}F_{h_{ij}}^{\prime\prime}(h_{ij}^t)h_{ij}^t
\end{eqnarray*}
Thus, $G(h_{ij}, h_{ij}^t)\geq F_{h_{ij}}$.

Replacing $G(h_{ij}, h_{ij}^t)$ in Eq.~(\ref{equ:argmin}) by Eq.~(\ref{equ:auxiliary}) and setting $\frac{\partial G(h_{ij}, h_{ij}^t)}{\partial h_{ij}}$ to be 0 result in the update rule in Eq.~(\ref{equ:h}) in the paper. 

Since Eq.~(\ref{equ:auxiliary}) is an auxiliary function, $F_{h_{ij}}$ is non-increasing under this update rule.

The update rule for $w_{ij}$ can be proved similarly.

\subsection{Complete Results on Comparison of NMF and NMF-Guttman in terms of ROC curve and Precision-Recall curve.}
The comparison of NMF and NMF-Guttman in terms of ROC curve and Precision-Recall curve on three MOOCs are shown in Figure~\ref{fig:roc}.
\begin{figure}[!htb]
    \centering
    \begin{subfigure}[t]{0.23\textwidth}
        \centering
        \includegraphics[scale=0.41]{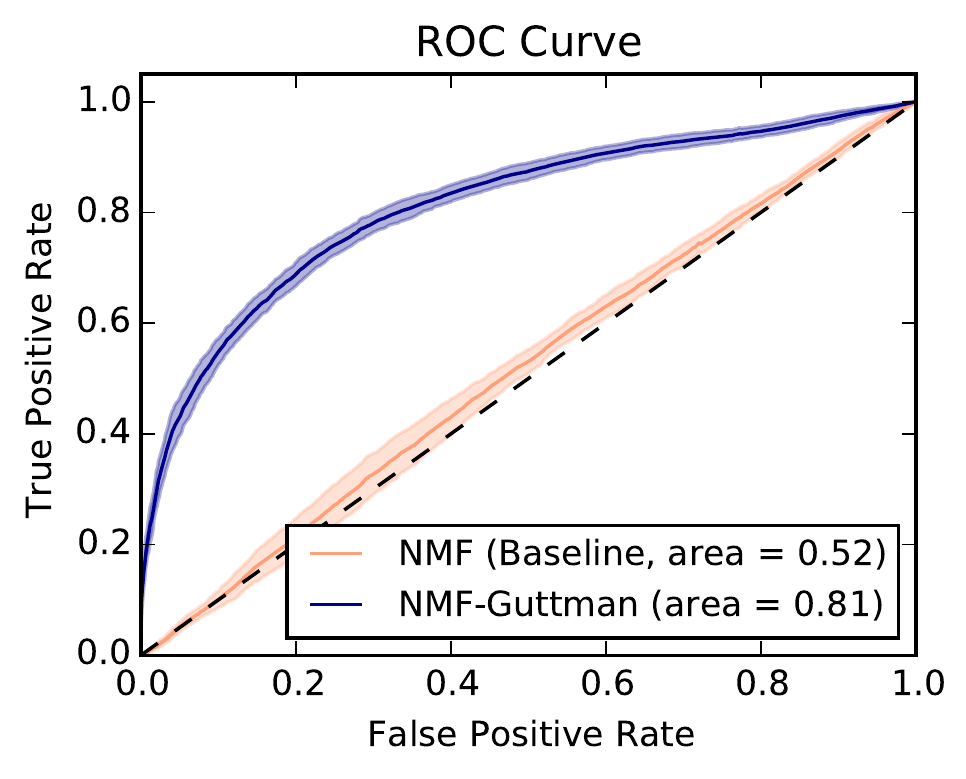}
        \caption{OPT}
    \end{subfigure}%
    ~ 
    \begin{subfigure}[t]{0.23\textwidth}
        \centering
        \includegraphics[scale=0.41]{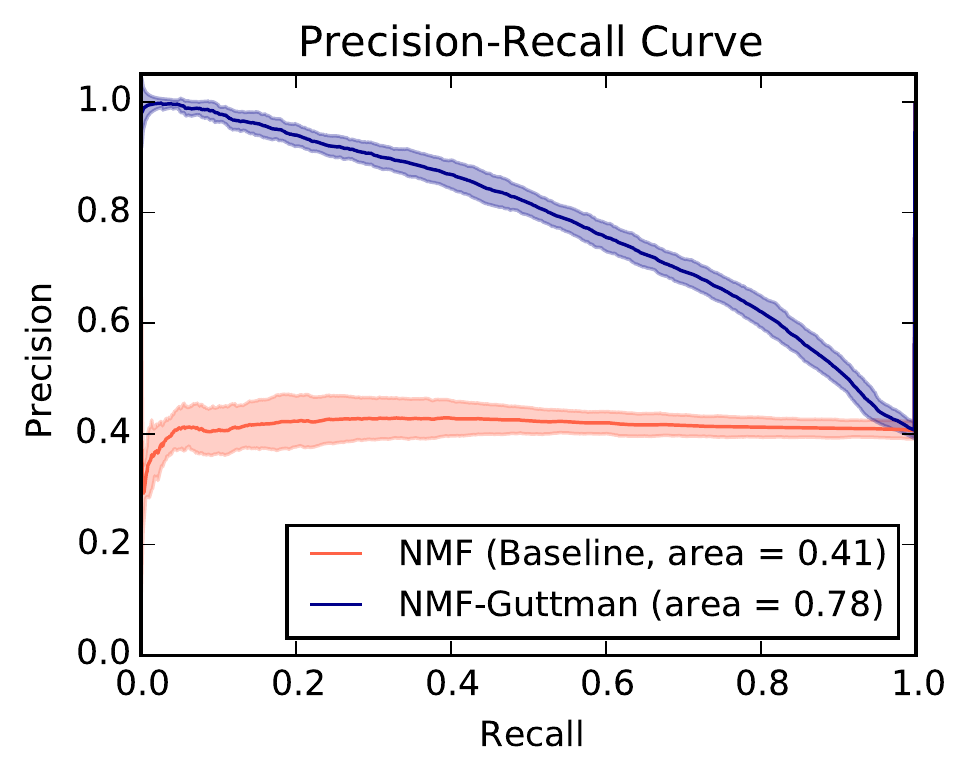}
        \caption{OPT}
    \end{subfigure}
    \begin{subfigure}[t]{0.23\textwidth}
        \centering
        \includegraphics[scale=0.41]{figures/postMACRO-ROC_scale_avg.pdf}
        \caption{ECON}
    \end{subfigure}%
    ~ 
    \begin{subfigure}[t]{0.23\textwidth}
        \centering
        \includegraphics[scale=0.41]{figures/postMACRO-PR_scale_avg.pdf}
        \caption{ECON}
    \end{subfigure}
    \begin{subfigure}[t]{0.23\textwidth}
        \centering
        \includegraphics[scale=0.41]{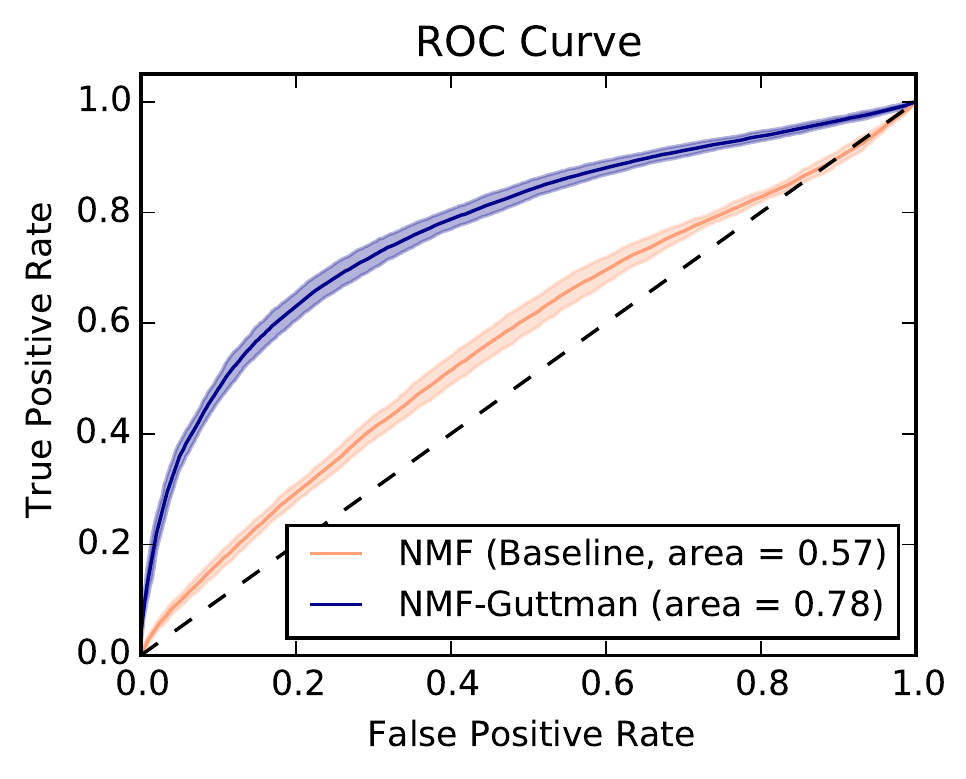}
        \caption{EDU}
    \end{subfigure}%
    ~ 
    \begin{subfigure}[t]{0.23\textwidth}
        \centering
        \includegraphics[scale=0.41]{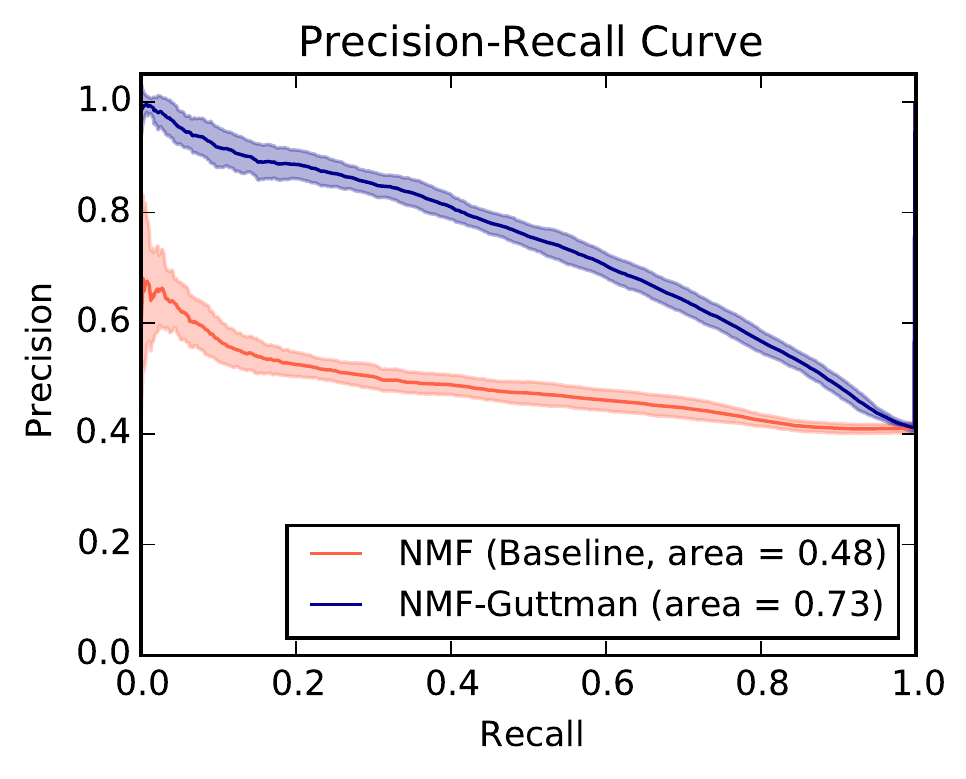}
        \caption{EDU}
    \end{subfigure}
    
    \caption{Comparison of NMF and NMF-Guttman in terms of ROC curve and Precision-Recall curve.}
    \label{fig:roc}
\end{figure}

\subsection{Experimental Results of Parameter Sensitivity on Regularization Parameter $\lambda_0$}
The performance of CR and $\norm{\vbf{V-WH}}$ with varying $\lambda_0$ are shown in Figure~\ref{fig:w0}. 
\begin{figure}[!htb]
    \centering	
    \begin{subfigure}[t]{0.23\textwidth}
        \centering
        \includegraphics[scale=0.30]{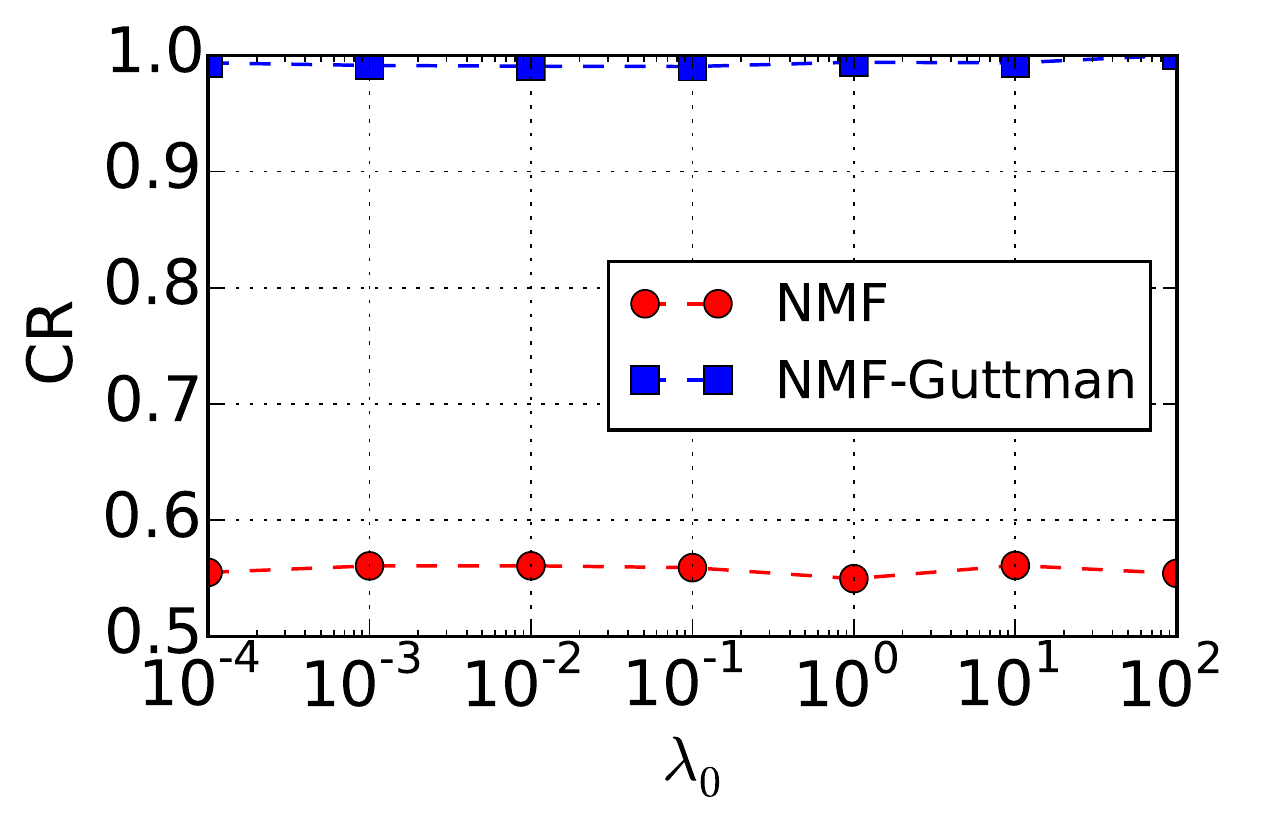}
        \caption{CR on OPT}
    \end{subfigure}%
    ~ 
    \begin{subfigure}[t]{0.23\textwidth}
        \centering
        \includegraphics[scale=0.30]{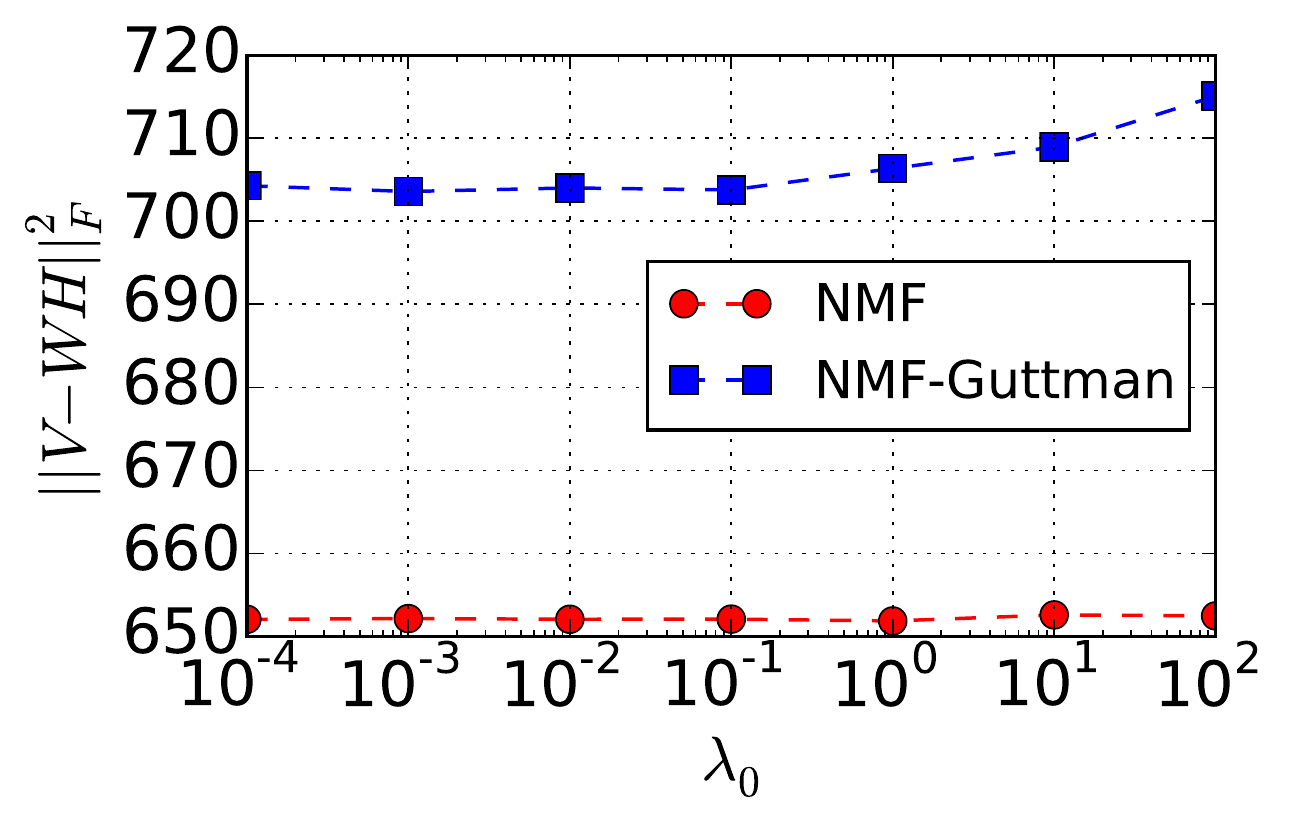}
        \caption{$\norm{\vbf{V-WH}}$ on OPT}
    \end{subfigure}
    \begin{subfigure}[t]{0.23\textwidth}
        \centering
        \includegraphics[scale=0.30]{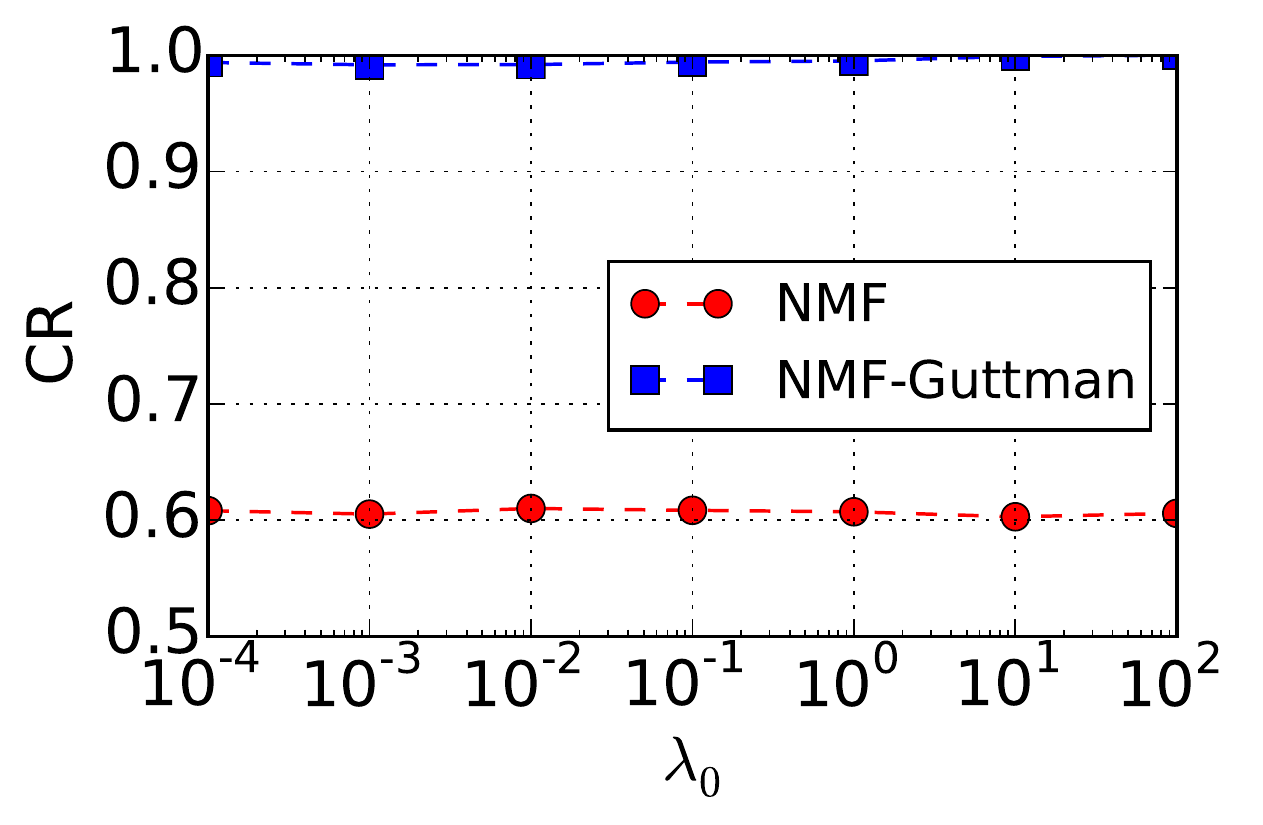}
        \caption{CR on ECON}
    \end{subfigure}%
    ~ 
    \begin{subfigure}[t]{0.23\textwidth}
        \centering
        \includegraphics[scale=0.30]{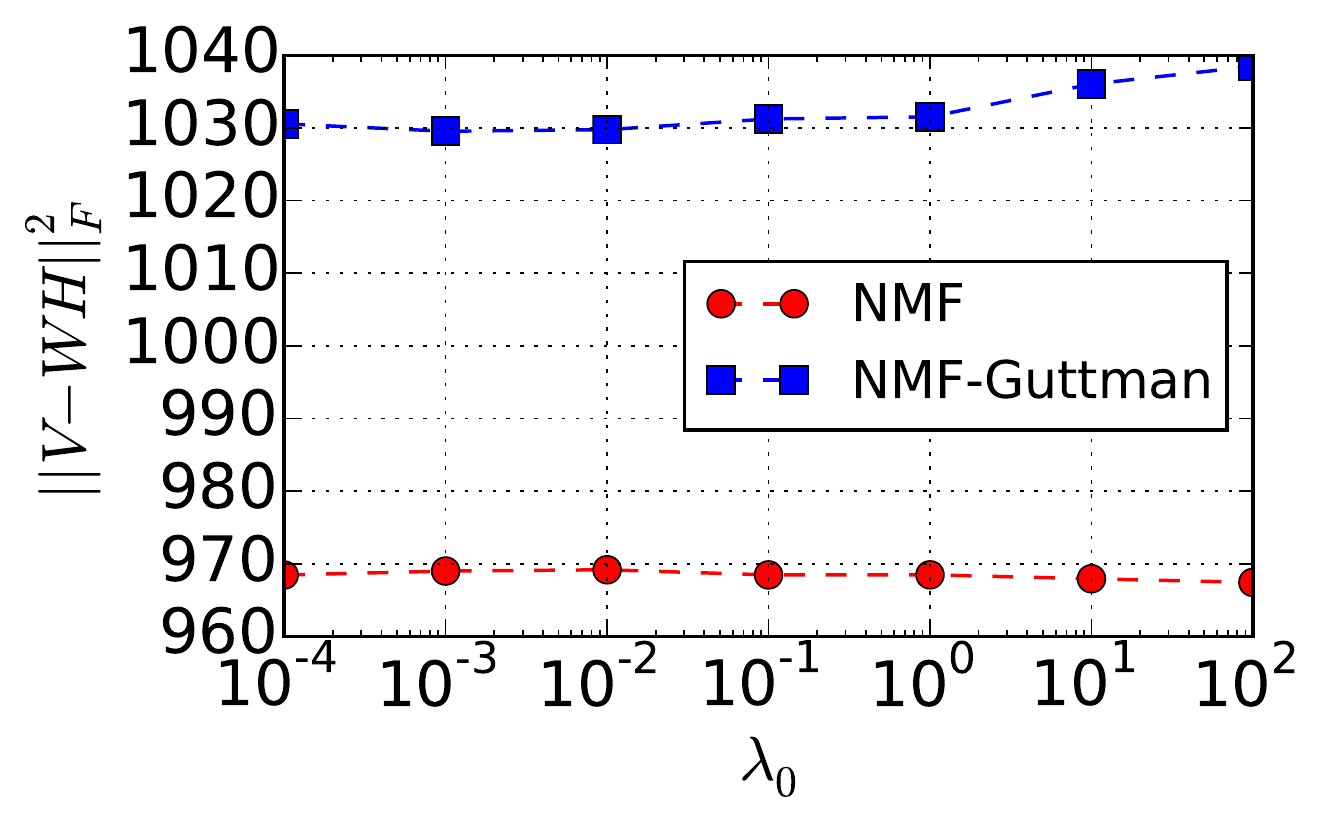}
        \caption{$\norm{\vbf{V-WH}}$ on ECON}
    \end{subfigure}
    
    \begin{subfigure}[t]{0.23\textwidth}
        \centering
        \includegraphics[scale=0.30]{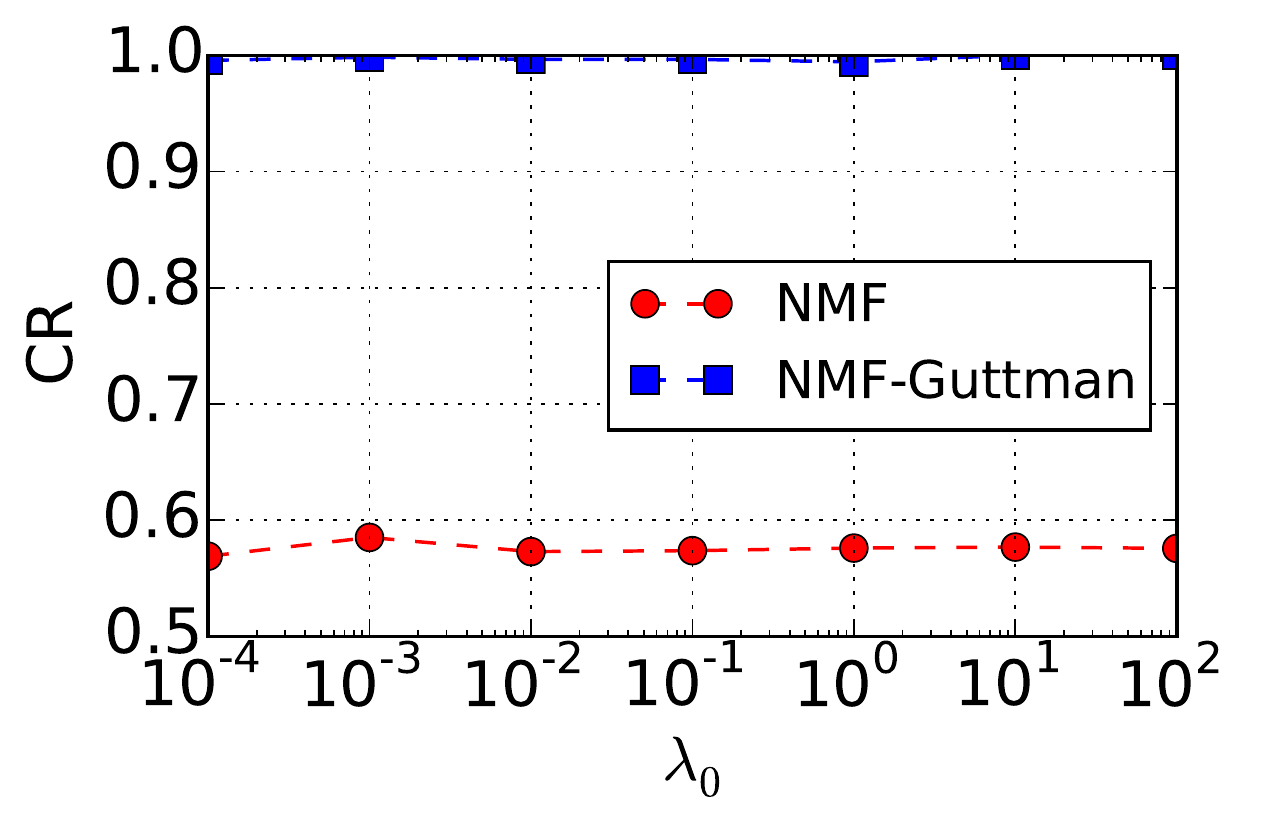}
        \caption{CR on EDU}
    \end{subfigure}%
    ~ 
    \begin{subfigure}[t]{0.23\textwidth}
        \centering
        \includegraphics[scale=0.30]{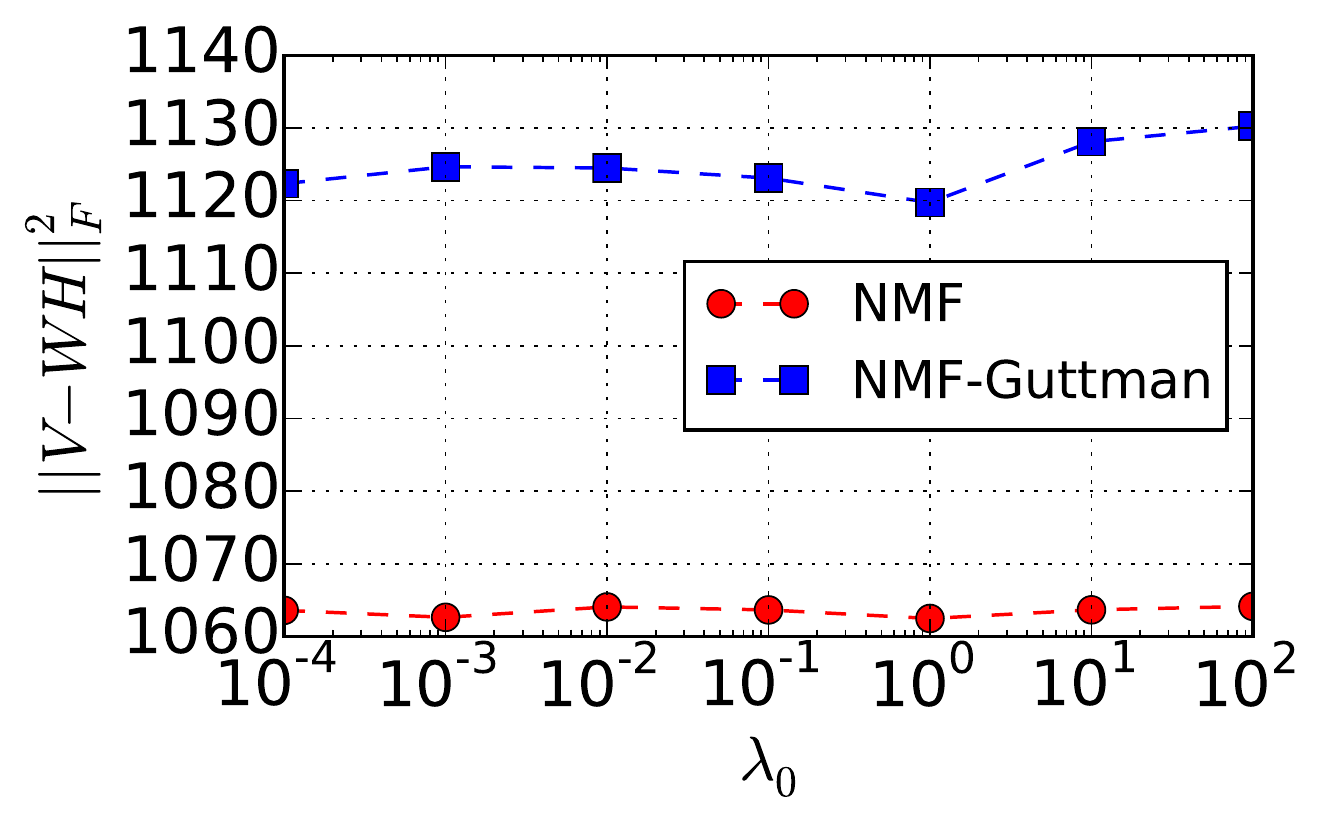}
        \caption{$\norm{\vbf{V-WH}}$ on EDU}
    \end{subfigure}
    \caption{Comparison of NMF and NMF-Guttman in terms of CR and $\norm{\vbf{V-WH}}$ with varying $\lambda_{0}$.}
    \label{fig:w0}
\end{figure}

\subsection{Complete Experimental Results of Parameter Sensitivity on Regularization Parameter $\lambda_1$}
The performance of CR and $\norm{\vbf{V-WH}}$ with varying $\lambda_1$ are shown in Figure~\ref{fig:hideal}. 
\begin{figure}[htb]
    \centering
    \begin{subfigure}[t]{0.23\textwidth}
        \centering
        \includegraphics[scale=0.30]{figures/do001_1_100_10-CRlog-hideal.pdf}
        \caption{CR on OPT}
    \end{subfigure}%
    ~ 
    \begin{subfigure}[t]{0.23\textwidth}
        \centering
        \includegraphics[scale=0.30]{figures/do001_1_100_10-objlog-hideal.pdf}
        \caption{$\norm{\vbf{V-WH}}$ on OPT}
    \end{subfigure}
    
    \begin{subfigure}[t]{0.23\textwidth}
        \centering
        \includegraphics[scale=0.30]{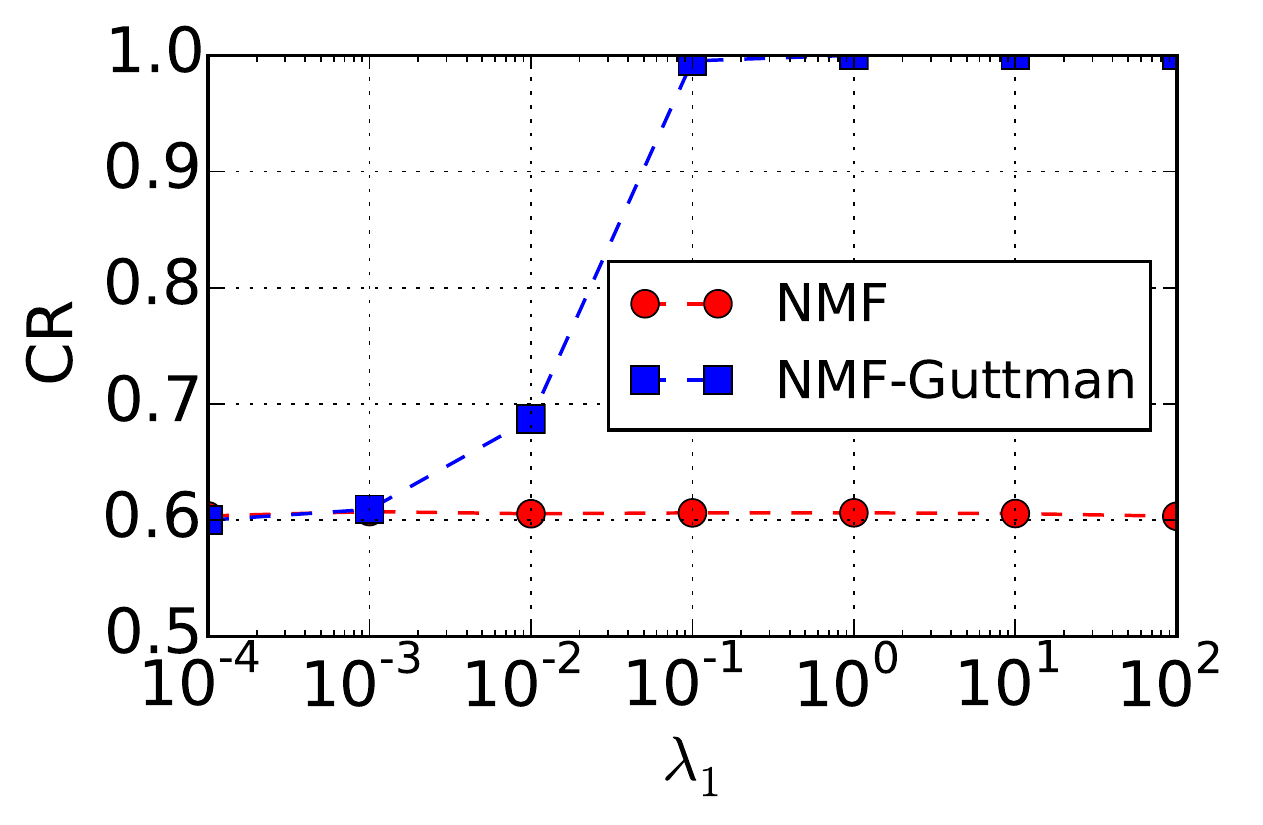}
        \caption{CR on ECON}
    \end{subfigure}%
    ~ 
    \begin{subfigure}[t]{0.23\textwidth}
        \centering
        \includegraphics[scale=0.30]{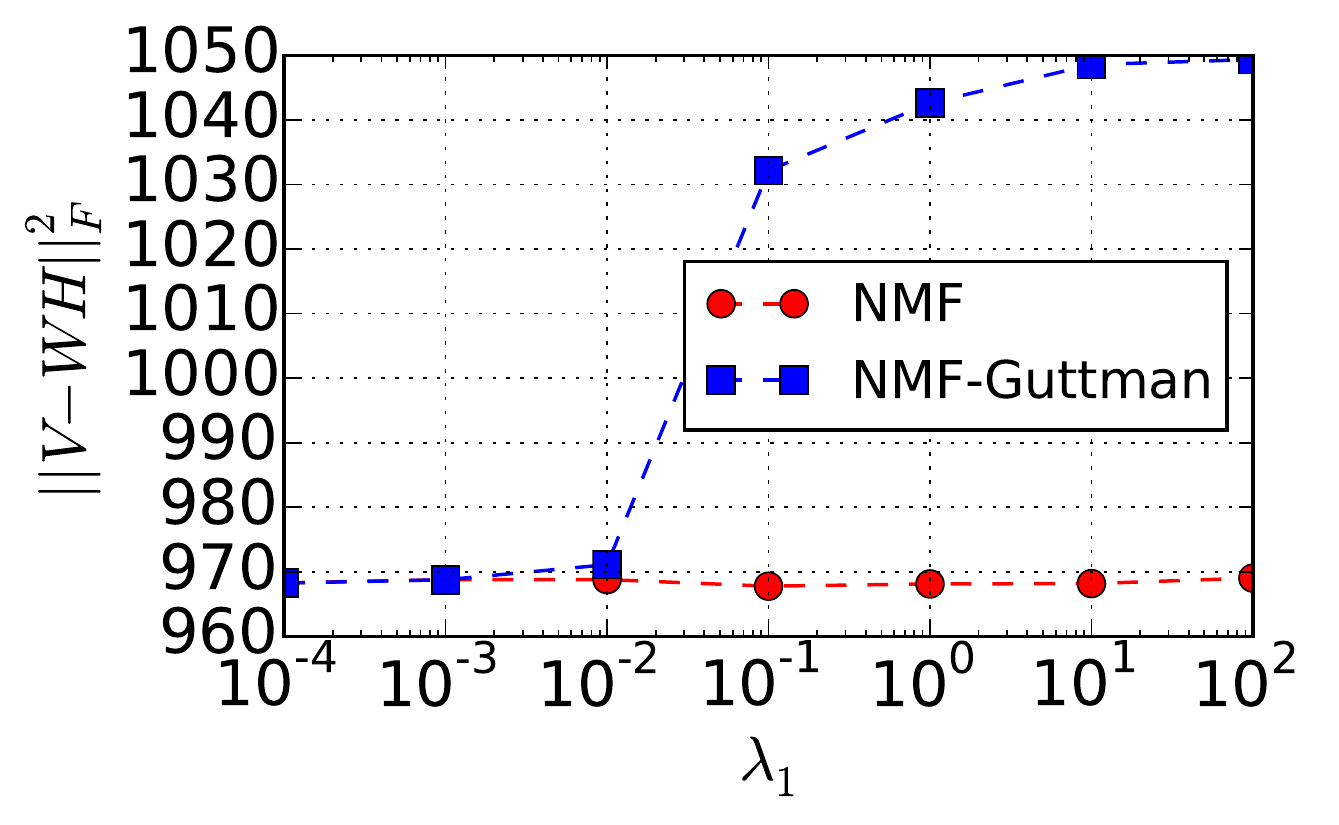}
        \caption{$\norm{\vbf{V-WH}}$ on ECON}
    \end{subfigure}
    
    \begin{subfigure}[t]{0.23\textwidth}
        \centering
        \includegraphics[scale=0.30]{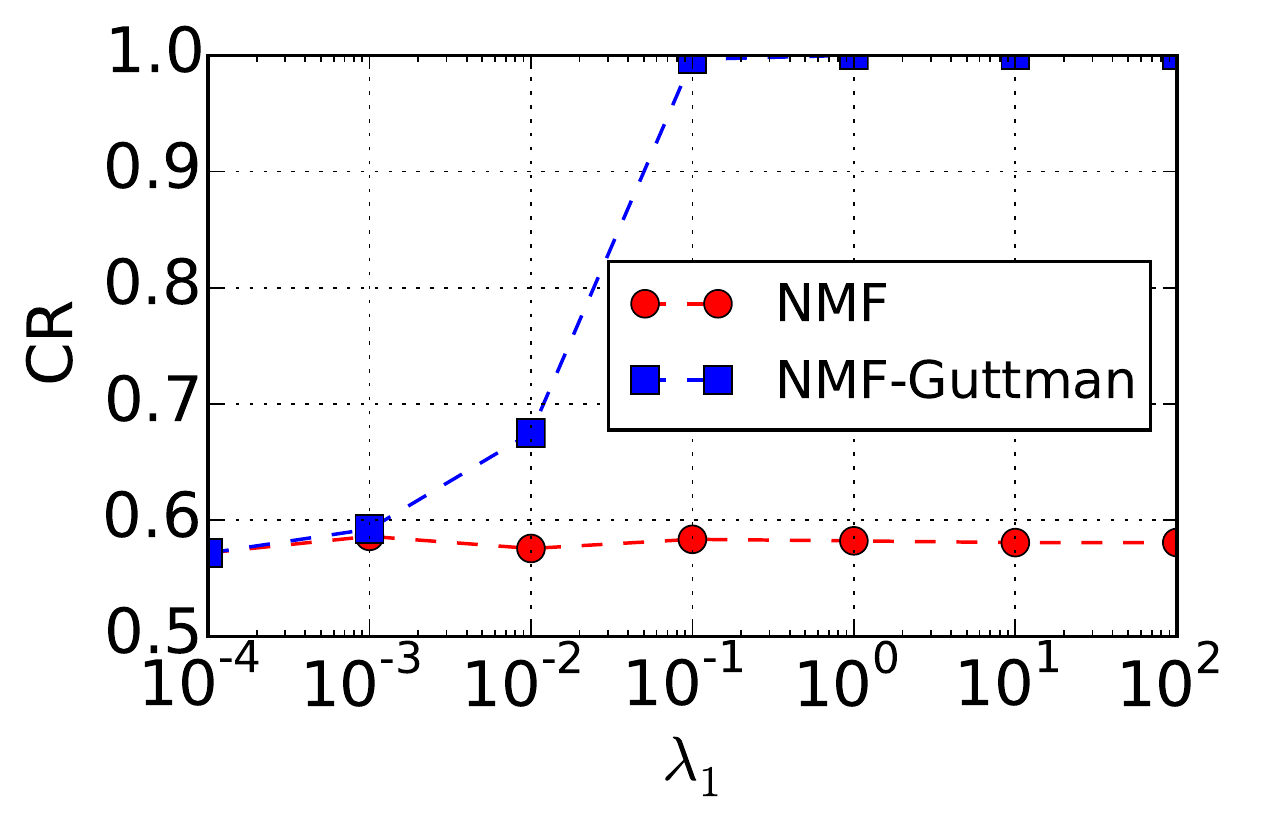}
        \caption{CR on EDU}
    \end{subfigure}%
    ~ 
    \begin{subfigure}[t]{0.23\textwidth}
        \centering
        \includegraphics[scale=0.30]{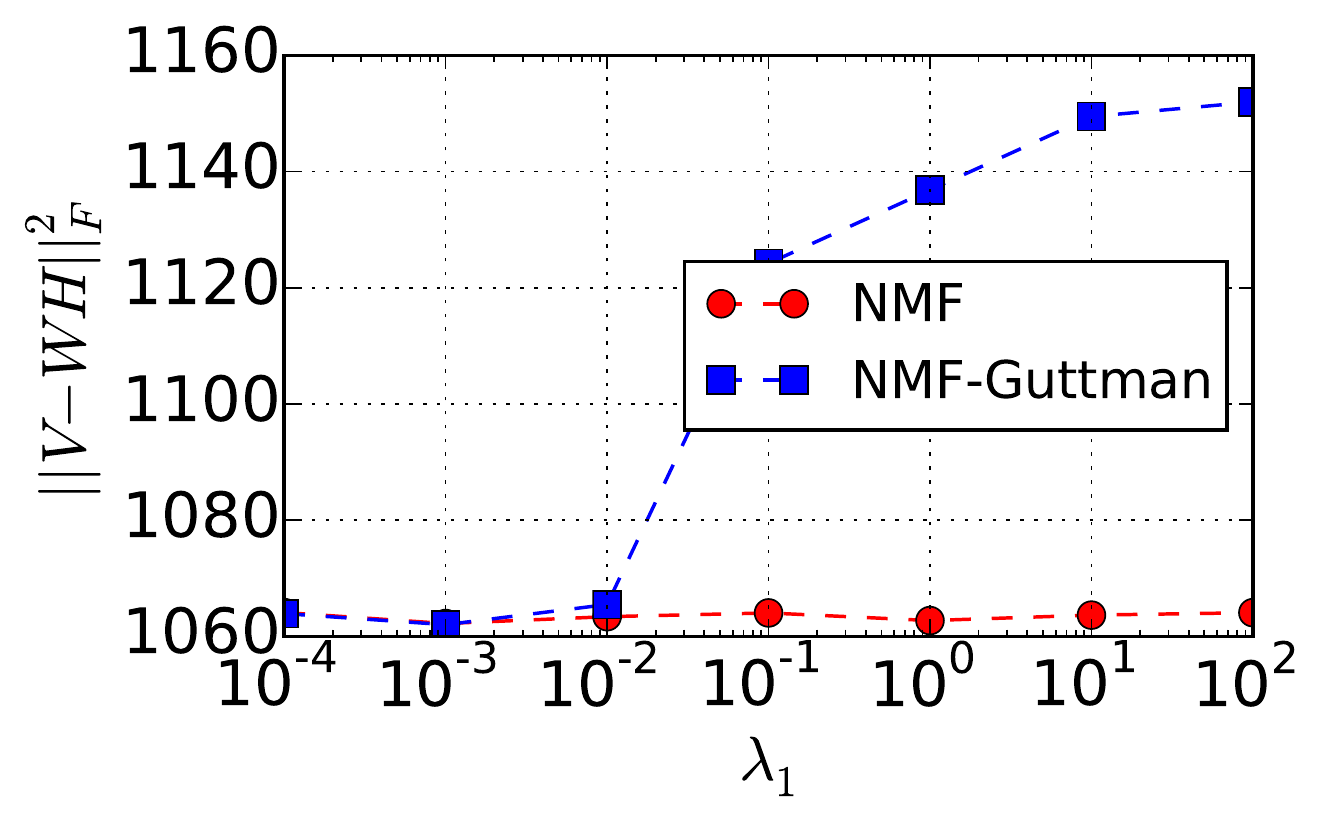}
        \caption{$\norm{\vbf{V-WH}}$ on EDU}
    \end{subfigure}
    \caption{Comparison of NMF and NMF-Guttman in terms of CR and $\norm{\vbf{V-WH}}$ with varying $\lambda_1$.}
    \label{fig:hideal}
\end{figure}

\subsection{Experimental Results of Parameter Sensitivity on Regularization Parameter $\lambda_2$}
The performance of CR and $\norm{\vbf{V-WH}}$ with varying $\lambda_2$ are shown in Figure~\ref{fig:hbinary}. 
\begin{figure}[!htb]
    \centering	
    \begin{subfigure}[t]{0.23\textwidth}
        \centering
        \includegraphics[scale=0.30]{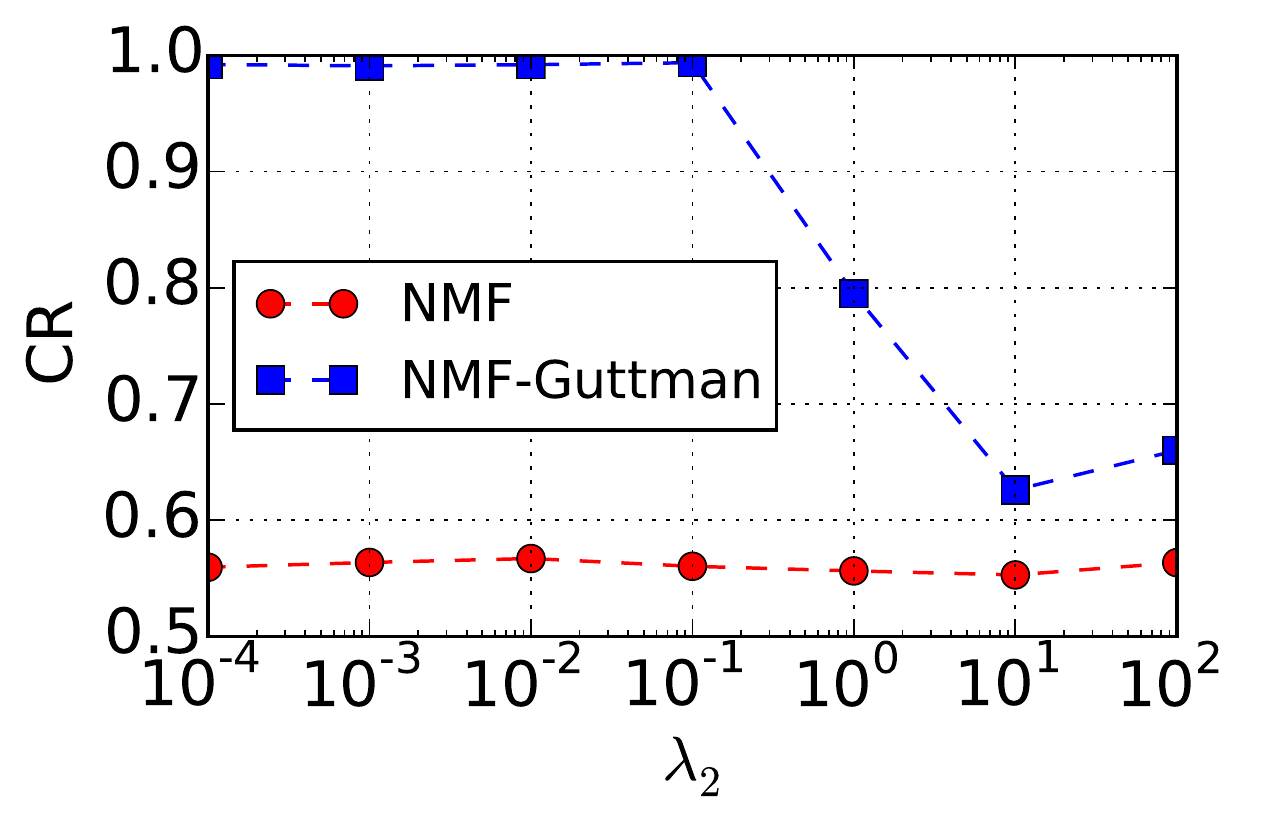}
        \caption{CR on OPT}
    \end{subfigure}%
    ~ 
    \begin{subfigure}[t]{0.23\textwidth}
        \centering
        \includegraphics[scale=0.30]{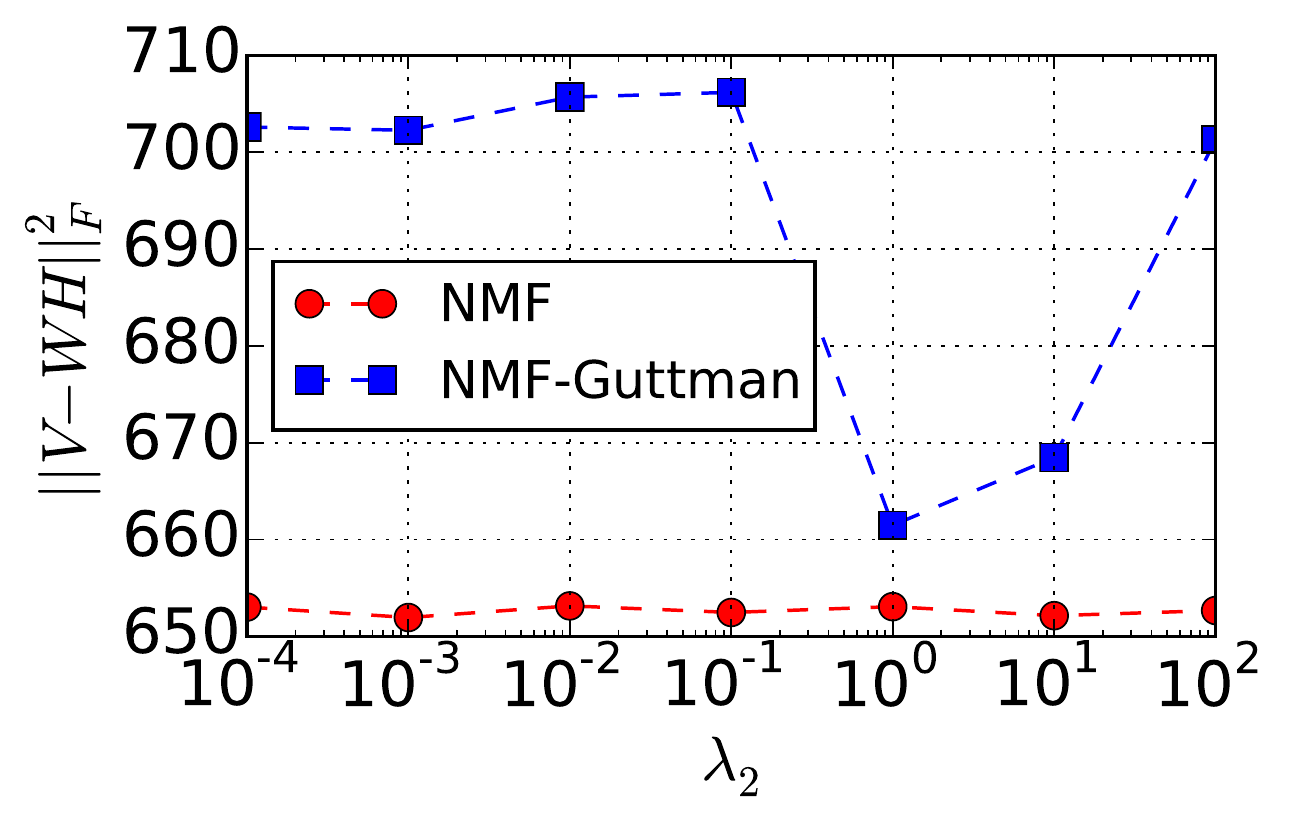}
        \caption{$\norm{\vbf{V-WH}}$ on OPT}
    \end{subfigure}
    \begin{subfigure}[t]{0.23\textwidth}
        \centering
        \includegraphics[scale=0.30]{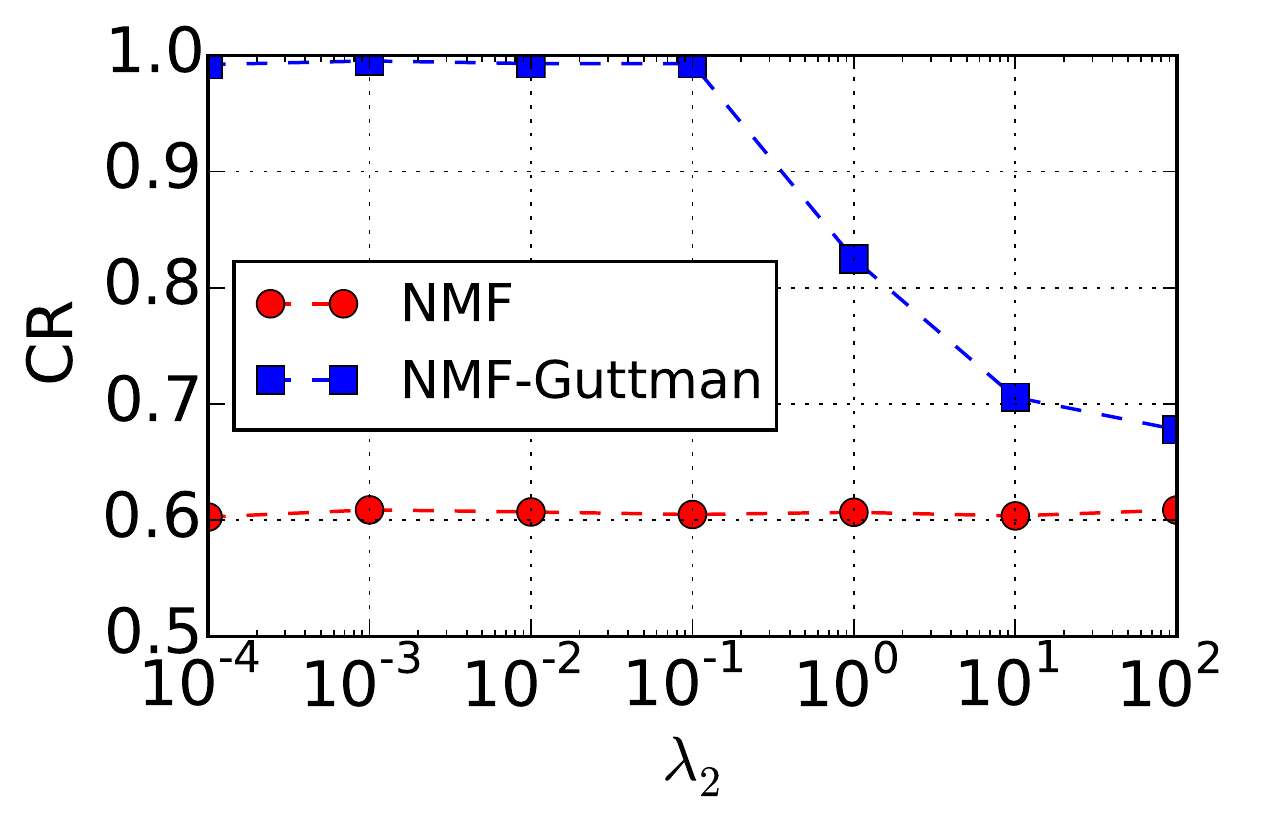}
        \caption{CR on ECON}
    \end{subfigure}%
    ~ 
    \begin{subfigure}[t]{0.23\textwidth}
        \centering
        \includegraphics[scale=0.30]{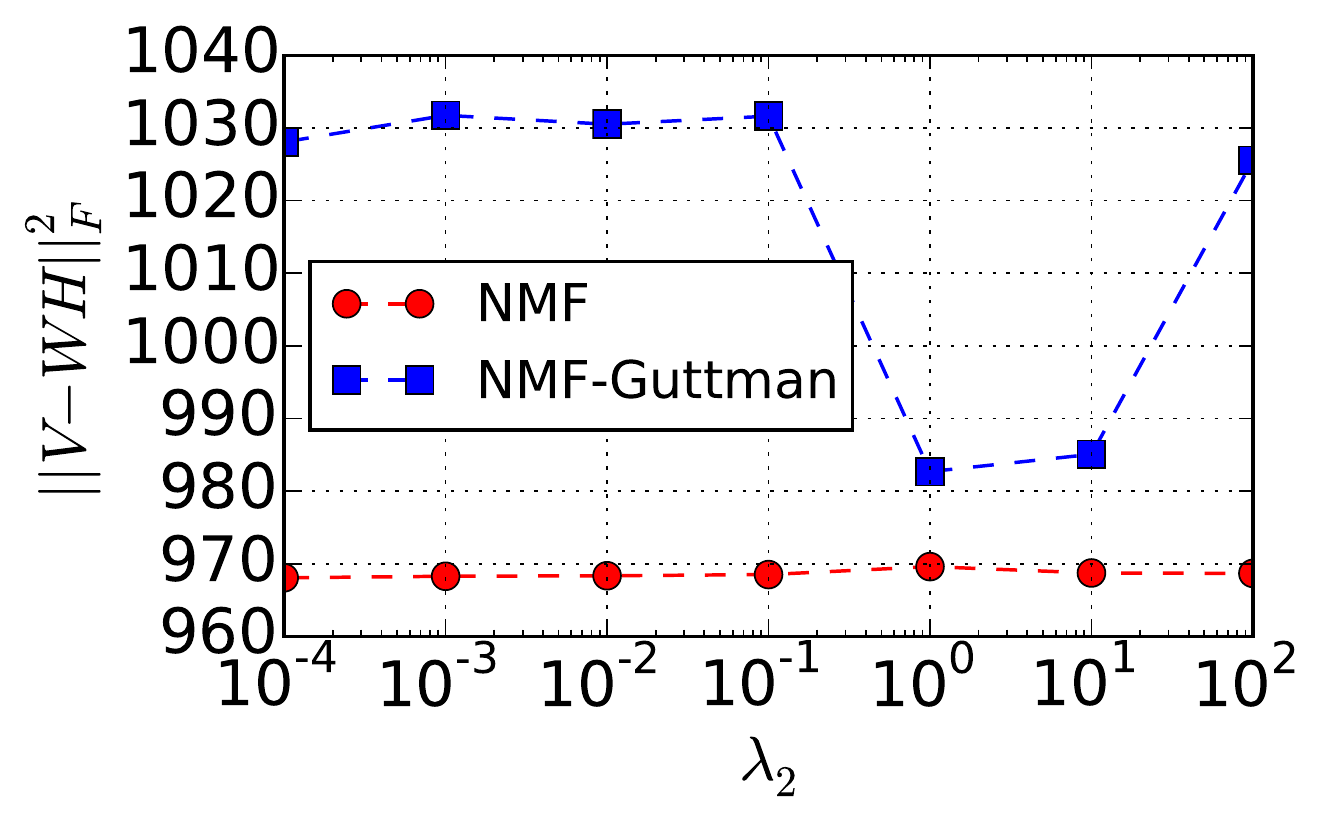}
        \caption{$\norm{\vbf{V-WH}}$ on ECON}
    \end{subfigure}
    
    \begin{subfigure}[t]{0.23\textwidth}
        \centering
        \includegraphics[scale=0.30]{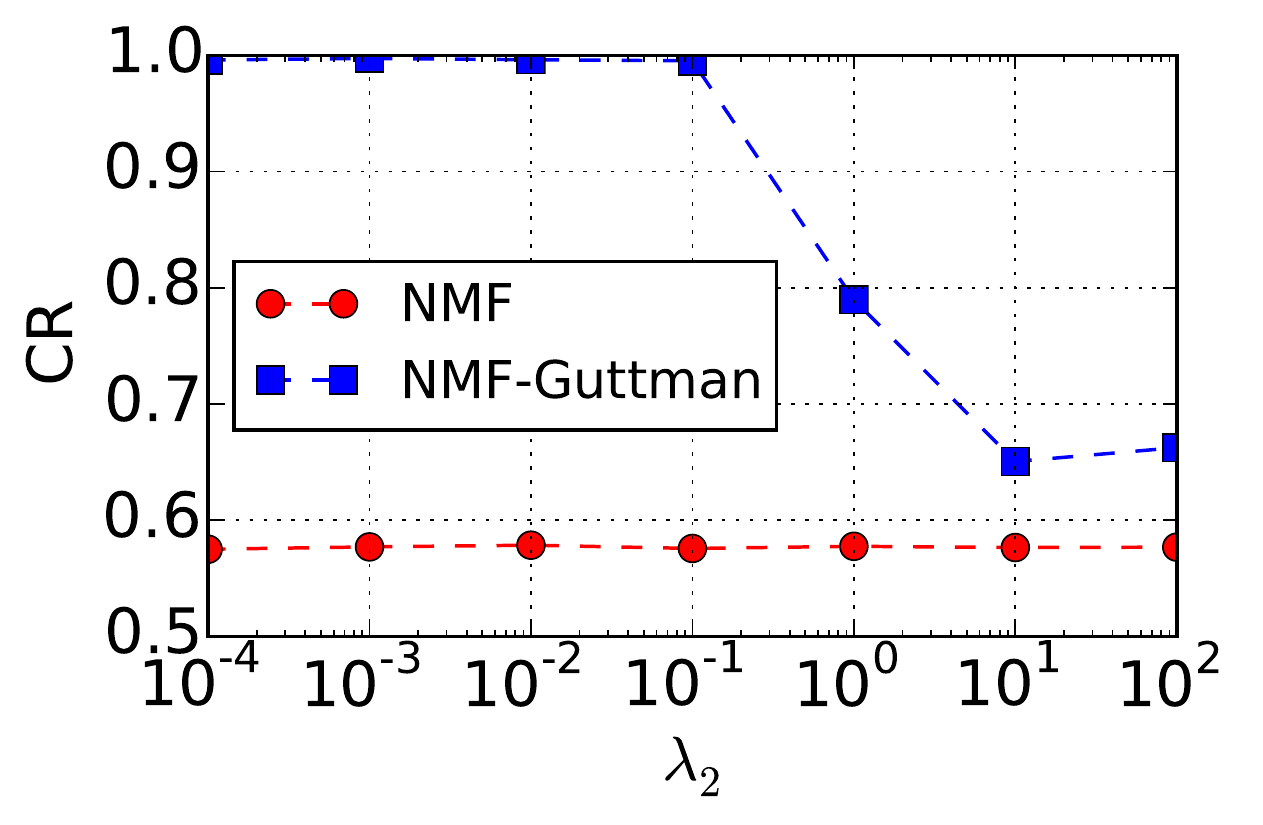}
        \caption{CR on EDU}
    \end{subfigure}%
    ~ 
    \begin{subfigure}[t]{0.23\textwidth}
        \centering
        \includegraphics[scale=0.30]{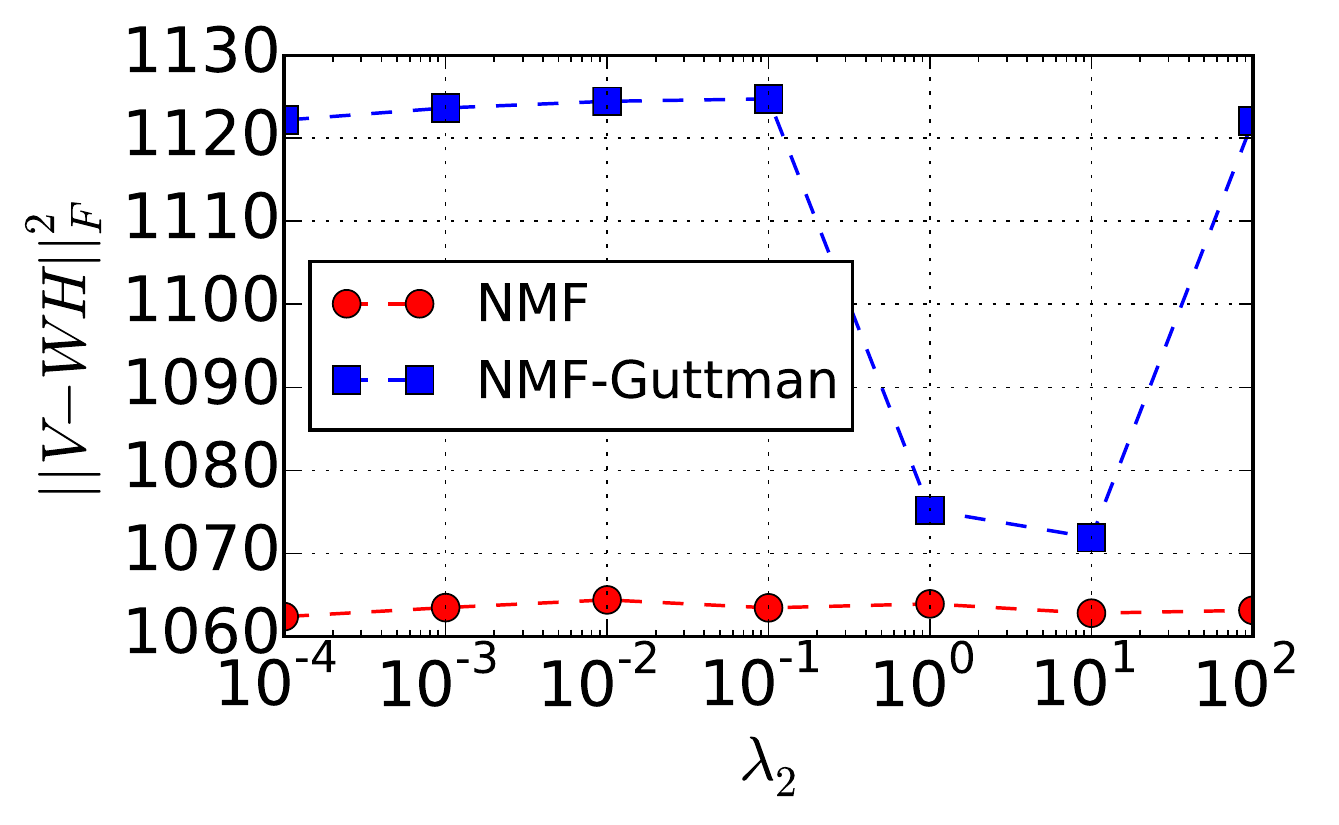}
        \caption{$\norm{\vbf{V-WH}}$ on EDU}
    \end{subfigure}
    \caption{Comparison of NMF and NMF-Guttman in terms of CR and $\norm{\vbf{V-WH}}$ with varying $\lambda_{2}$.}
    \label{fig:hbinary}
\end{figure}

\subsection{Experimental Results of Parameter Sensitivity on Regularization Parameter $k$}
The performance of CR and $\norm{\vbf{V-WH}}$ with varying $k$ are shown in Figure~\ref{fig:topic}. 
\begin{figure}[!htb]
    \centering	
    \begin{subfigure}[t]{0.23\textwidth}
        \centering
        \includegraphics[scale=0.30]{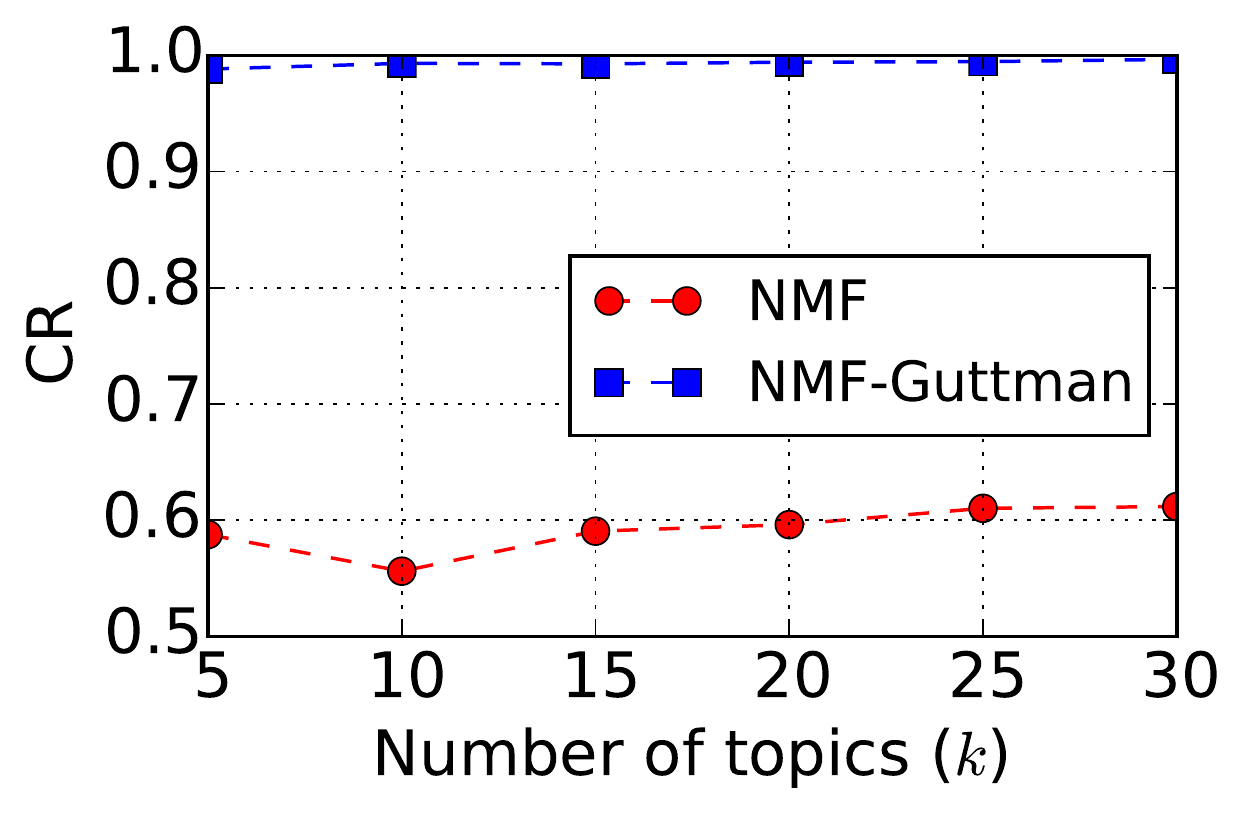}
        \caption{CR on OPT}
    \end{subfigure}%
    ~ 
    \begin{subfigure}[t]{0.23\textwidth}
        \centering
        \includegraphics[scale=0.30]{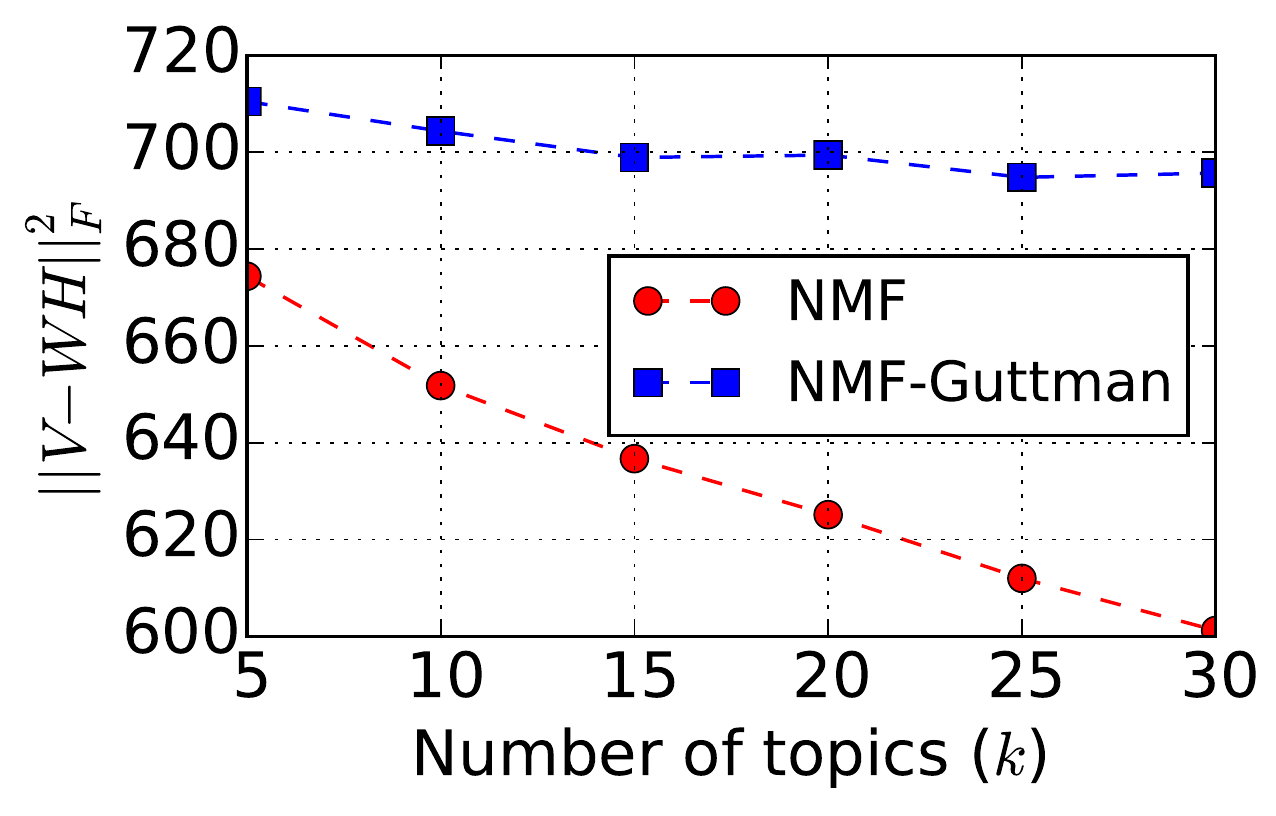}
        \caption{$\norm{\vbf{V-WH}}$ on OPT}
    \end{subfigure}
    \begin{subfigure}[t]{0.23\textwidth}
        \centering
        \includegraphics[scale=0.30]{figures/do001_1_1_30-CRlogrank.pdf}
        \caption{CR on ECON}
    \end{subfigure}%
    ~ 
    \begin{subfigure}[t]{0.23\textwidth}
        \centering
        \includegraphics[scale=0.30]{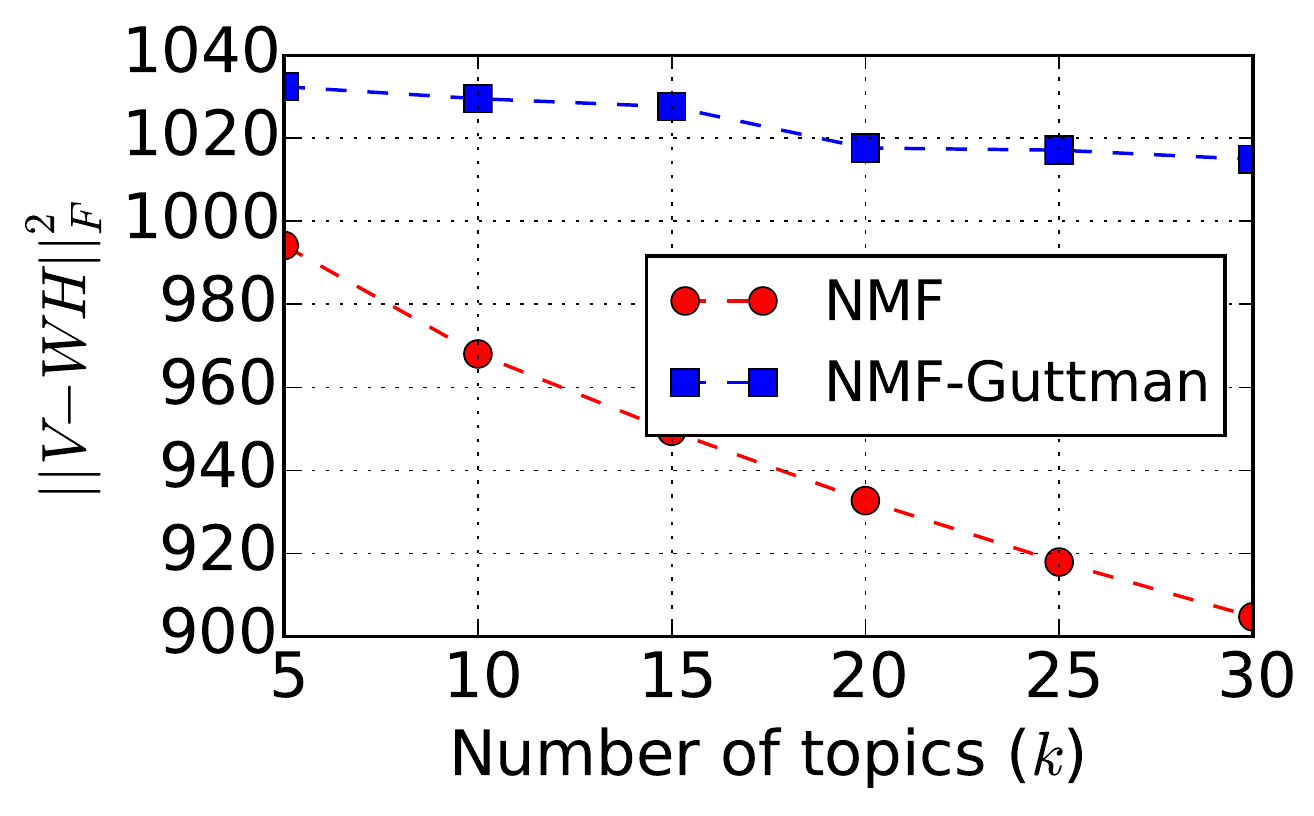}
        \caption{$\norm{\vbf{V-WH}}$ on ECON}
    \end{subfigure}
    
    \begin{subfigure}[t]{0.23\textwidth}
        \centering
        \includegraphics[scale=0.30]{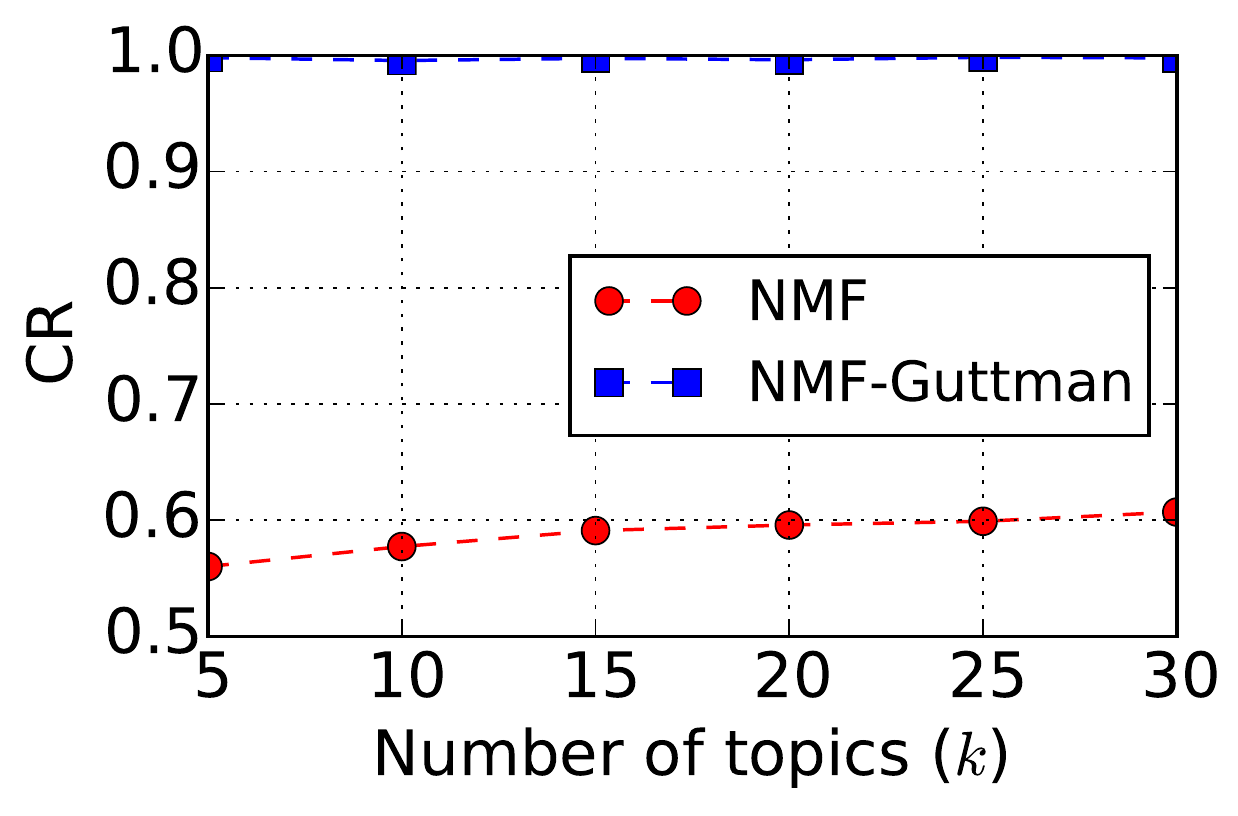}
        \caption{CR on EDU}
    \end{subfigure}%
    ~ 
    \begin{subfigure}[t]{0.23\textwidth}
        \centering
        \includegraphics[scale=0.30]{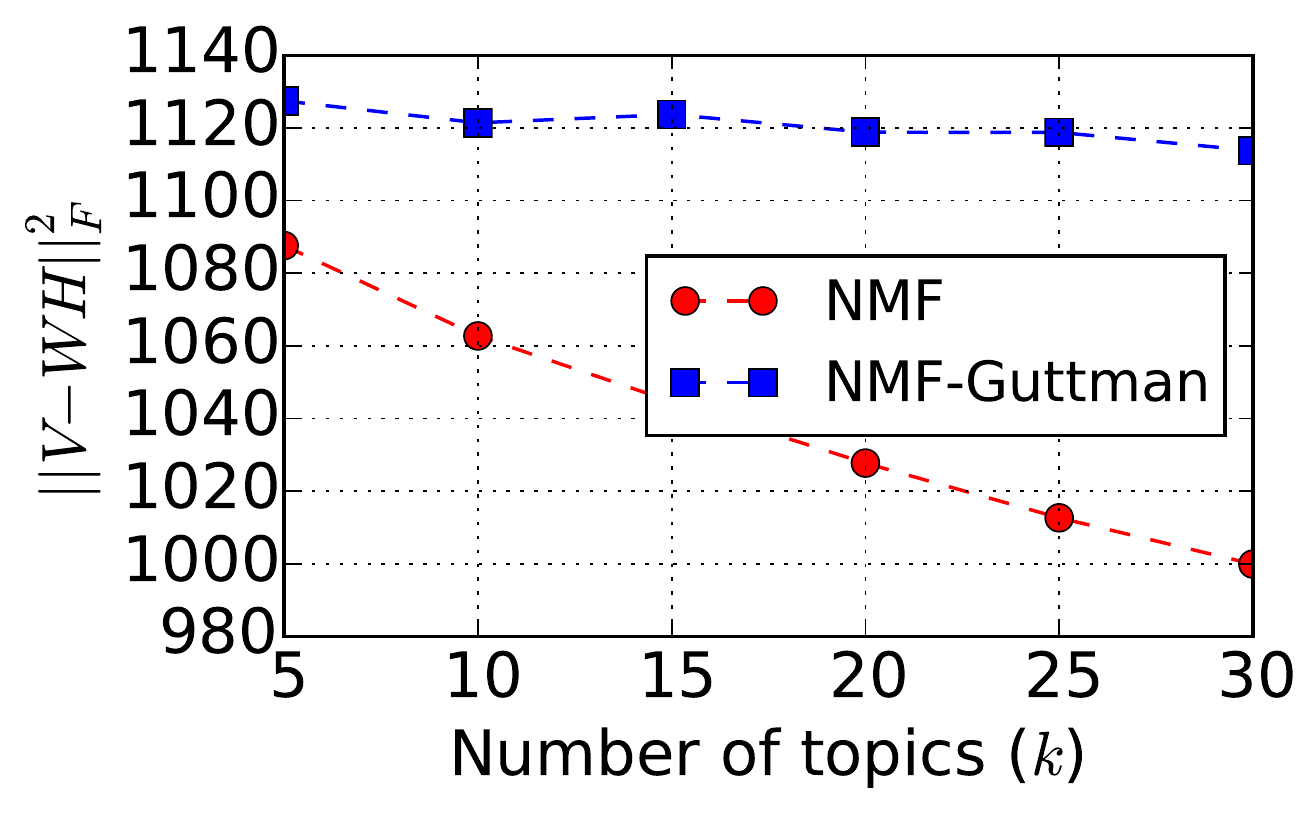}
        \caption{$\norm{\vbf{V-WH}}$ on EDU}
    \end{subfigure}
    \caption{Comparison of NMF and NMF-Guttman in terms of CR and $\norm{\vbf{V-WH}}$ with varying the number of topics ($k$).}
    \label{fig:topic}
\end{figure}

\subsection{Interpretations on OPT MOOC Topics Generated from NMF and NMF-Guttman}
Table~\ref{tab:nmfint} and Table~\ref{tab:nmfguttmanint} show Interviewee 1's interpretation on OPT MOOC topics generated from NMF and NMF-Guttman. 

\begin{table*}
    \centering
    \small
    \caption{Interviewee 1's interpretation on OPT MOOC topics generated from NMF.}
    \begin{tabular}{cp{7.2cm}l}
        \toprule
        No. & Topics  & Interpretation \\
        \toprule
        
        1 & course thank really learn would like assign time lecture great & Thanks! \\
        
        2 & video lecture load 001 optimization chrome detail coursera org class & Platform\\
        
        3& file pi line urlib2 lib submit python27 data solver req & Python external solvers\\
        
        4 & submit assignment assignment\_id view message screen namehttp 3brows detail 001 & Platform/submission\\
        
        5 & problem knapsack solution value optimization submit grade thank solve get & How to submit assignments\\
        
        6 & python solver use matlab command install pyc run java window & Python/Java/Matlab and extend solver (How to start)\\
        
        7 & item value weight capacity estimate take node knapsack tree calculation & Dynamic programing for knapsack, how to understand and code\\
        
        8 & color node graph order clique number use iteration degree edge & Graph coloring, how to use and understand graph theory concepts\\
        
        9 & solution opt problem use search get custom move time optimize &  Traveling salesman problem, trying to improve algorithm/customise \\
        
        10 & use dp memory column bb algorithm bound table implement time & Comparing algorithms memory/time\\
        \bottomrule
    \end{tabular}%
    \label{tab:nmfint}%
\end{table*}%

\begin{table*}
    \small
    \centering
    \caption{Interviewee 1's interpretation on OPT MOOC topics generated from NMF-Guttman with inferred difficulty ranking.}
    \begin{tabular}{cp{4.8cm}lc} %>{\em}
        \toprule
        No. & Topics  & Interpretation &  Inferred Ranking \\
        \toprule
        1 & python problem file solver assign pi class video course use & How to use platform/python & 1\\
        
        2 & submit thank please pyc grade feedback run solution check object & Platform/submission issues & 2\\
        
        3 & warehouse one result edge exactly list decide tour lib suppose & How to read\&use data for facility location & 3\\
        
        4 & solution optimize best first much insert move want feasible less & How to improve/create heuristics for knapsack problems & 4\\
        
        5 & color opt random search local greedy swap node good get & Understand and implement local search & 5\\
        
        6 & point mip certificate puzzle enough le route model course de & Course structure (eg. what's enough to get certificate?) & 6 \\
        
        7 & use scip two try implement lp differ need solver easy & How to implement LP/MIP and solver availability & 7\\
        
        8 & time temperature sa move opt would like well start ls & How to design and tune simulated annealing and local search & 8 \\
        
        9 & problem warehouse custom 10 tsp cluster mip constraint vehicle solution & Relationship between problems and algorithms & 9\\
        
        10 & item use value node solution problem algorithm optimize time dp & Knapsack (using dynamic programing), how to speed up algorithms & 10\\
        \bottomrule
    \end{tabular}%
    \label{tab:nmfguttmanint}%
\end{table*}%

\subsection{Interpretations on EDU MOOC Topics Generated from NMF and NMF-Guttman}
Table~\ref{tab:nmfedu} and Table~\ref{tab:guttmanEDU} show the course coordinator's interpretation on EDU MOOC topics generated from NMF and NMF-Guttman. 

\begin{table*}
    \centering
    \small
    \caption{The course coordinator's interpretation on EDU MOOC topics generated from NMF.}
    \begin{tabular}{cp{8cm}p{8cm}}
        \toprule
        No. & Topics  & Interpretation \\
        \toprule
        
        1 & student cp think use would skill teacher need task assess & Discussion of assesemnt of collaborative problems solving discussed in week 2 of the course\\
        
        2 & teach course hello teacher english hope everyone name hi improve & Welcome introductions to the course\\
        
        3& assign peer evaluate grade thank course score mooc mark assess & Discussion about peer assessemtns in the course \\
        
        4 & skill century 21st assess develop learn curriculum need interest education & General discussion of introducory ideas in the course \\
        
        5 & org 001 atc21s http coursera 971791 human\_grading assessments courses class & General discussion about the approach to assessemtns in the course\\
        
        6 & problem solve collaborate idea skill think differ group task cp & Discussion about the nature of collaborative problems solving, week 2 of the course\\
        
        7 & learn forward look student assess hi excit collaborate everyone course & Introductory comments about the course\\
        
        8 & technology learn use education teacher new learner change us way & Discussion about the impact of technology on the curriculum\\
        
        9 & school education year australia current primary hi interest melbourne name & Talk between participants about their background \\ 
        
        10 & work thank group really help together routine know time student & Thank you notes and discussion at the end of the course\\
        \bottomrule
    \end{tabular}%
    \label{tab:nmfedu}%
\end{table*}%

\begin{table*}
    \small
    \centering
    \caption{The course coordinator's interpretation on EDU MOOC topics generated from NMF-Guttman with inferred difficulty ranking.}
    \begin{tabular}{cp{6cm}p{7.45cm}c} %p{7.3cm}
        \toprule
        No. & Topics  & Interpretation &  Inferred Ranking \\
        \toprule
        1 & learn student teach teacher skill course assess use school collaboration  & Discussion of relationship between teaching and learning as in week 1 of the course syllabus & 1\\
        
        2 & forward name also philippin hi join india teach help better  & Establishing social presence: Introduction posts, and statements of and what people want to get out of the course & 2\\
        
        3 & assign peer thank one evaluate mark grade comment score could & Discussion about the peer assignments in the course & 3\\
        
        4 & skill develop need assess australia plan base progress social approach  & Discussion of developmental teaching and assessment as per 1st and 2nd week of course syllabus  & 4\\
        
        5 & 001 coursera org atc21s 971791 human\_grading  courses assessments submissions class & General discussion about course  process and structure & 5\\
        
        6 & cp task problem think collaborate solve differ activity group idea & Discussion of collaborative problem solving, which is the focus of week 2 syllabus  & 6 \\
        
        7 & student assess individual australia provide report less level progress may & Discussion of student level, individualised reporting against performance levels, focus of weeks 2 and 3 of syllabus & 7\\
        
        8 & use make great give many way thank bring agree human &  Appreciation posts at the end of the course & 8 \\
        
        9 & would school level week link read model table set observe & Unclear & 9\\
        
        10 & work student group thank one need time make together know & Discussion of difference between collaboration and group-work, theme through the course, and in assignments  & 10\\
        \bottomrule
    \end{tabular}%
    \label{tab:guttmanEDU}%
\end{table*}%

}{
}

\end{document}